\documentclass[11pt]{article}

\usepackage[preprint]{acl}

\usepackage{times}
\usepackage{latexsym}

\usepackage[T1]{fontenc}

\usepackage[utf8]{inputenc}

\usepackage{microtype}

\usepackage{inconsolata}

\usepackage{graphicx}

\usepackage{amssymb}
\usepackage{dsfont}
\usepackage{amsmath}
\usepackage{booktabs}
\usepackage{array}
\usepackage{tabularx}
\usepackage{makecell}
\usepackage{arydshln}
\usepackage{multirow}

\usepackage{todonotes}
\usepackage{ulem}
\usepackage{enumitem}

\usepackage{xcolor}

\usepackage[most]{tcolorbox}
\tcbuselibrary{listings,breakable,skins}

\providecommand{\method}{CPS}

\title{AdaTutor: Adaptive Tutoring through Pedagogical Strategy Selection}

\author{Rajeev Bhatt Ambati \\
  UNC Chapel Hill \\
  \texttt{ambati@cs.unc.edu} \\\And
  Shashank Srivastava \\
  UNC Chapel Hill \\
  \texttt{ssrivastava@cs.unc.edu} \\\And
  Snigdha Chaturvedi \\
  UNC Chapel Hill \\
  \texttt{snigdha@cs.unc.edu}}

\begin{document}
\maketitle
\begin{abstract}
Effective tutoring requires determining not only \textit{what} information to provide, but also \textit{which form} of assistance to deliver. A tutor may address the same student error using various pedagogical strategies, adapting its approach as the interaction progresses. We introduce \textit{Comparative Pedagogical Supervision} (\method{}), a training framework to learn to select the optimal strategy and feedback for student learning. We evaluate \method{} on math and science reasoning datasets. After five tutoring turns, \method{} outperforms the strongest fixed-strategy baselines. The learned tutor transfers across datasets, and its strategy selection can also inform a separate API-based tutor.
\end{abstract}

\begin{figure}[!htb]
  \centering
  \includegraphics[width=\columnwidth]{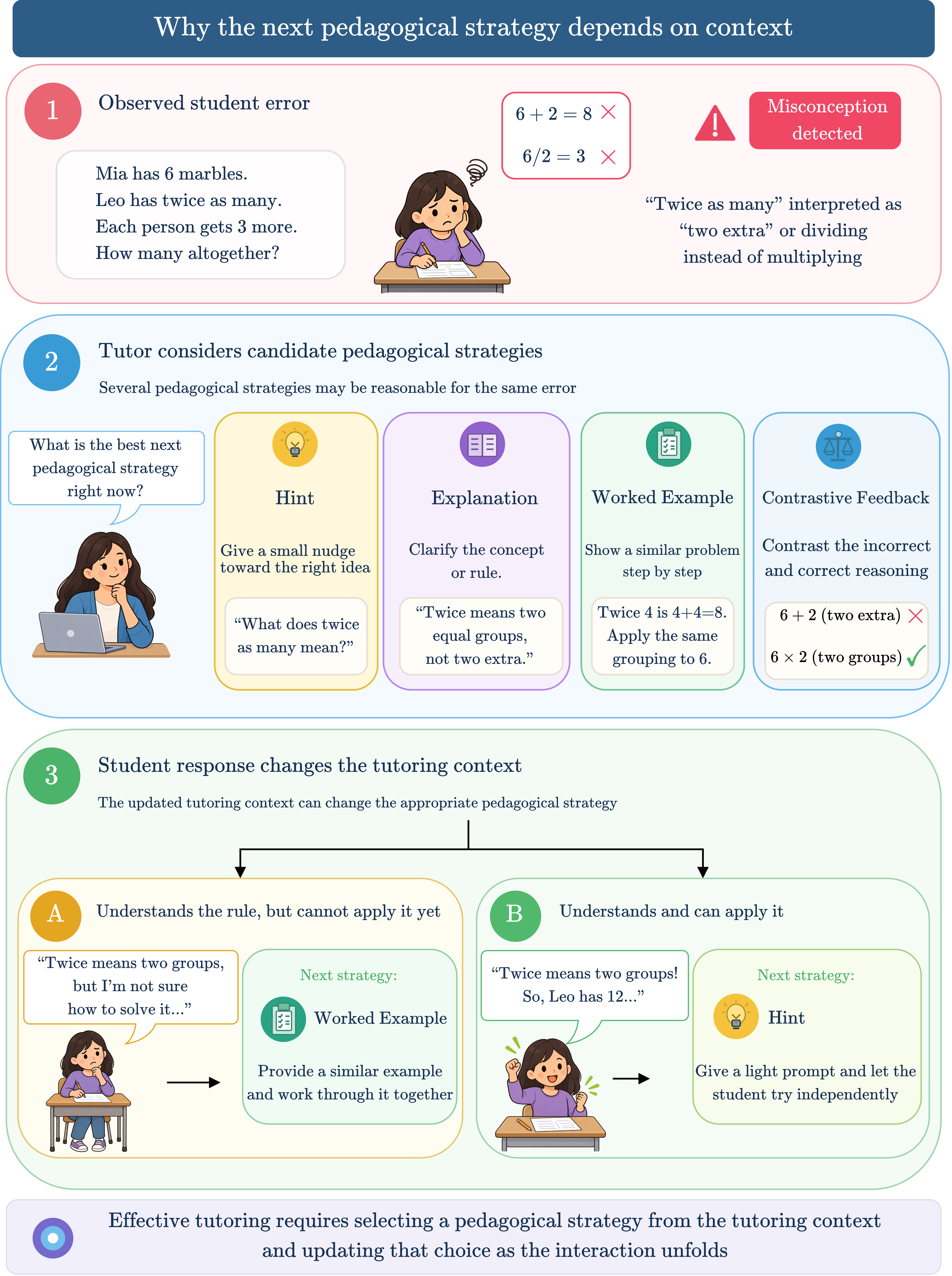}
  \caption{
  \textbf{Pedagogical strategy varies according to the tutoring context:} Depending on the student's response, further support may be provided through a \textsc{Worked example} or a \textsc{Hint} for subsequent steps.
  }
  \label{fig:intro}
\end{figure}

\section{Introduction}
Intelligent Tutoring Systems (ITS) must balance instructional support with opportunities for independent problem solving, a challenge known as the \textit{assistance dilemma} \citep{koedinger2007}. Research in cognitive science shows that the level of support needed varies as students progress \citep{salden2009}, and identifies distinct forms of assistance such as hints, scaffolding, worked examples, and self-explanation \citep{vanlehn2006,wood1976,chi1989}. Tutors built on Large Language Models (LLMs) must therefore decide not only what information to provide, but also which form of assistance suits a student's misconception and progress.

For example, Figure~\ref{fig:intro} depicts a student who initially misinterprets the phrase ''twice as many.'' As the student's responses reveal progress or continuing difficulties, the tutor should reconsider which pedagogical strategy to use next. A student who understands the relationship but cannot apply it may benefit from a \textsc{Worked example} illustrating the concept, whereas a student who applies it correctly may require only a \textsc{Hint} for the remaining steps.

LLMs make it easier to generate various forms of pedagogical assistance. Recent studies use pedagogical dialogues and error-grounded supervision \citep{mathdial2023,sonkaralignment2024,daheim2024}. Other approaches train tutors to provide effective guidance \citep{scarlatos2025,tutorrl2025} and learn to select tutoring actions or prompts \citep{nam2025efficient,changprompt2026}. The appropriate strategy depends on the student's current work and how the interaction has progressed. Our \textit{key idea} is to \textit{adapt the pedagogical strategy} at each turn based on the student's progress and remaining difficulties.



To train AdaTutor, we introduce Comparative Pedagogical Supervision (\method{}). At each tutoring context, the tutor generates its response conditioned on a jointly identified pedagogical strategy. We use six pedagogical strategies grounded in cognitive science research: \textsc{Hint}, \textsc{Step scaffold}, \textsc{Explicit explanation}, \textsc{Worked example}, \textsc{Self-explanation}, and \textsc{Contrastive feedback} (Section~\ref{sec:method_details}). To train the tutor, we use an LLM to generate one response for each of the six strategies. We then use independent simulated student samples to estimate the probability of solving the problem following each response. The strategy-response pairs with the highest probability are then used to supervise the tutor. It generates a \texttt{[strategy] response} sequence in one call. The strategy tag makes this choice interpretable, and supervision eliminates the need for candidate selection during inference.


On the GSM8K, SVAMP, and ScienceQA datasets, \method{} improves final accuracy by 6.3, 2.6, and 10.1 percentage points, respectively, compared to the strongest fixed-strategy tutors after five tutoring turns. Subsequent analyses investigate the specific contributions of dialogue context, strategy selection, and response generation. Additionally, \method{} demonstrates transferability across datasets. On a GSM8K subset, integrating its learned strategy selection with GPT-4.1-mini improves final accuracy by 7.3 percentage points over the strongest fixed-strategy API tutors.

Our contributions are:
\begin{itemize}[topsep=1pt, leftmargin=*, noitemsep]
    \item We introduce \method{}, a framework to train AdaTutor that adapts its pedagogical strategy to the student's progress and difficulties.
    \item We demonstrate improved tutoring across three datasets and multiple students, with analyses of dialogue context, strategy selection, and response generation.
    \item We further demonstrate cross-dataset transfer and show that strategy selection learned by \method{} improves the performance of an independent API tutor.
\end{itemize}

\section{Previous Work}
\paragraph{Pedagogical Feedback and Remediation:} Pedagogical dialogue datasets and training objectives facilitate the generation of instructional responses. MathDial provides annotations of mathematical tutoring dialogues with teacher moves \citep{mathdial2023}. Pedagogical Alignment promotes scaffolded guidance using preference-based training \citep{sonkaralignment2024}. SocraticLM is trained on Socratic dialogues that involve six distinct student types \citep{socraticlm2024}. For error remediation, \citet{daheim2024} ground responses in stepwise verification of student reasoning. AdaTutor instead compares responses from each strategy based on their impact on subsequent student outcomes.

\paragraph{Training Tutoring Policies:} \citet{scarlatos2025} evaluate candidate responses based on predicted student correctness and a pedagogical rubric, subsequently training models with DPO. TutorRL employs multi-turn RL to balance student task success with pedagogical support \citep{tutorrl2025}. \citet{nam2025efficient} develop long-horizon policies for high-level tutoring actions derived from compact student representations. Learning to Prompt selects session-level pedagogical prompts using topic and subject information \citep{changprompt2026}, while LEAP generates scaffolding plans from inferred cognitive states \citep{leap2026}.

\method{} constructs supervision using alternative feedback from each strategy within the same tutoring context. The optimal strategy–feedback pairs are then used to train a tutor that jointly generates a strategy and feedback at each turn.

\section{Method}
\label{sec:method}

\begin{figure*}[!htb]
  \centering
  \includegraphics[width=\textwidth]{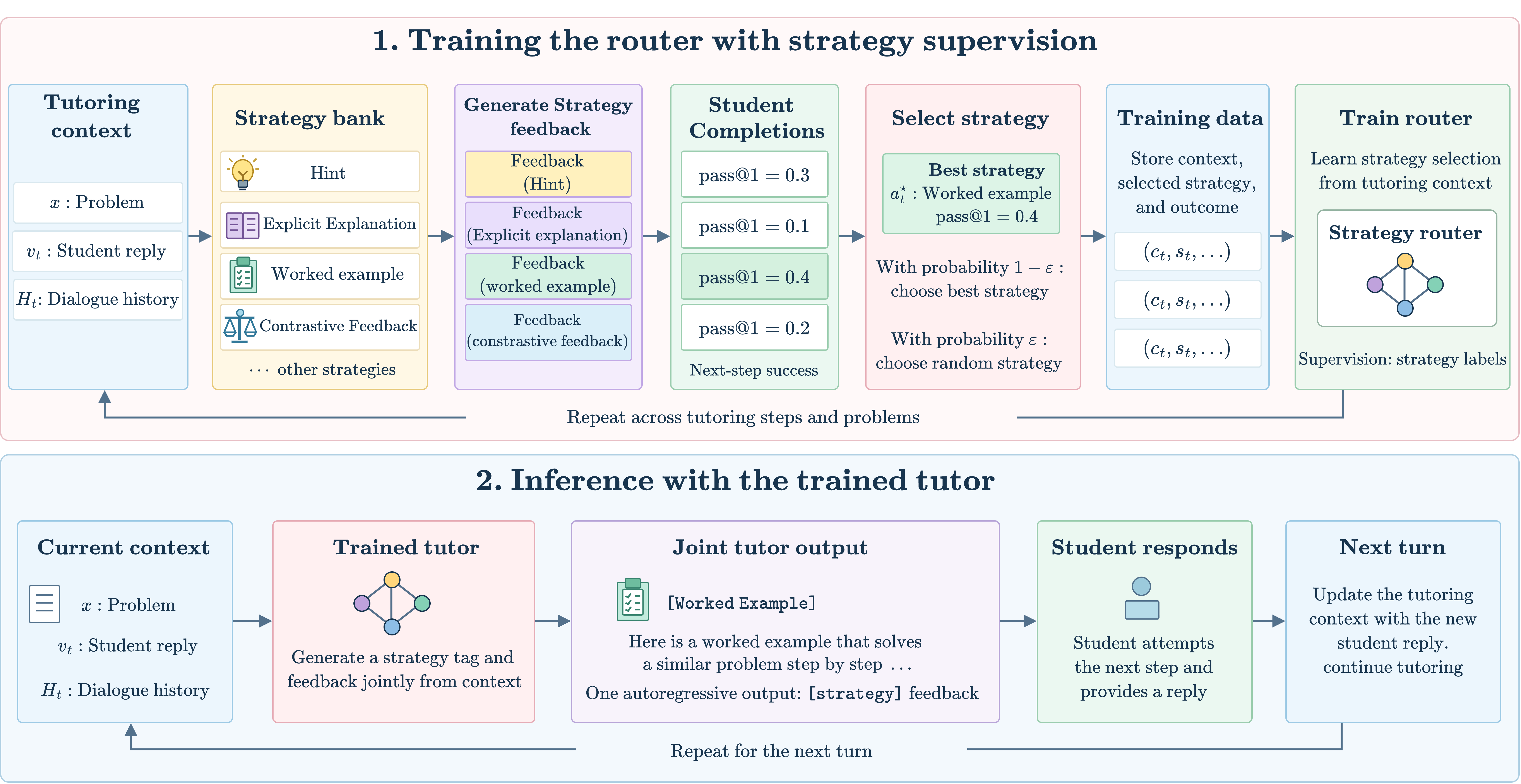}
  \caption{\textbf{Overview of \method{}:} During training, we sample one feedback for each pedagogical strategy from the tutor. Independent student attempts measure how often each strategy leads to a correct answer. A strategy and its feedback provide supervision when its score is positive and strictly higher than others. A separate exploration policy selects the strategy used to continue the multi-turn interaction. At inference, \method{} generates a '\texttt{[strategy] feedback}' in one call. strategy tag helps with interpretability.}
  \label{fig:main}
\end{figure*}

At each tutoring turn, \method{} selects a pedagogical strategy based on the student's work and the preceding exchanges. We train these choices by comparing feedback from different strategies at the same point in the dialogue. The student's answers after each feedback message determine which strategy and feedback to use as a training target (Figure~\ref{fig:main}).

\subsection{Problem Setup} 
Let $x$ denote a problem, $r$ its full reference solution with intermediate steps, and $y$ its final reference answer. Before tutor turn $t$, the public dialogue history is $H_t$. The tutor receives the \emph{tutoring context} $c_t=(x,r,H_t)$ and produces feedback $u_t$. The student receives the problem, public dialogue, and tutor feedback. It then produces a public reply $v_t$ and a separate answer to the original problem $x$. We check that answer against $y$. Each multi-turn interaction ends when the student answers correctly or reaches a budget of $T$ tutor turns.


\subsection{Pedagogical Strategies}
\label{sec:strategies}
Pedagogical research has long studied how different forms of tutoring assistance affect student learning. They differ in how much of the reasoning the tutor provides. A \textsc{Minimal hint} directs attention to the relevant concept and leaves the next step to the student \citep{vanlehn2006,aleven2003}. A \textsc{Step scaffold} breaks the next step into smaller actions \citep{wood1976}. An \textsc{Explicit explanation} states the relevant rule and the reasoning behind it \citep{wittwer2008}. A \textsc{Worked example} demonstrates the same reasoning on an analogous problem \citep{salden2009}. \textsc{Self-explanation} asks the student to justify their own reasoning \citep{chi1989}. \textsc{Contrastive feedback} compares the student's incorrect reasoning with a correct alternative \citep{daheim2024}. These strategies span low and high levels of assistance, which reflects the assistance dilemma \citep{koedinger2007}. They also address different sources of misconceptions. Appendix~\ref{app:strategies} provides definitions and examples.



\subsection{Comparative Pedagogical Supervision}
\label{sec:method_details}

\paragraph{Strategy generation and scoring:}
Let $\mathcal{A}$ be the set of six pedagogical strategies described in the previous section. Let $\mathcal{M}_0$ denote the base tutor model before our supervised training. At each tutoring context $c_t$ where the problem remains unsolved, we prompt $\mathcal{M}_0$ to generate one feedback $u_t^{(a)}$ for each strategy $a\in\mathcal{A}$. For each feedback $u_t^{(a)}$, we sample $K$ independent student attempts at the original problem $x$. Every attempt starts with the same student and public dialogue $H_t$, followed by that strategy's feedback. The score $q_t(a)$ for strategy $a$ is the fraction of these attempts with a correct answer(\textit{pass@1}). E.g., seven correct answers among ten attempts give a score of $0.7$. These attempts are used only to compare the strategies.

\paragraph{Supervised targets:} We retain a training example when one strategy has a positive score strictly higher than every other strategy. Let $a_t^\star$ denote this strategy. The training input is $c_t$, and the target contains the strategy tag $a_t^\star$ followed by its feedback $u_t^{(a_t^\star)}$. The target therefore has the format \texttt{[strategy] feedback}. The strategy tag helps with interpretability.



\paragraph{Multi-turn Supervision:} To train adaptation over multiple turns, we continue the interactions to collect examples from different strategies. We choose the highest-scoring strategy with probability $1-\varepsilon$, and with probability $\varepsilon$, we randomly choose from other strategies. We deliver its feedback $u_t$ to the student and obtain a fresh reply $v_t$. We append both messages to $H_t$ to form $H_{t+1}$, then compare strategies again at the updated context $c_{t+1}$. The supervised target still uses the highest-scoring strategy, even when a different strategy continues the multi-turn interaction. This separates the choice of training target from the choice of which tutoring situations to encounter next. Appendix~\ref{app:recipe} examines target selection, exploration, and the number of student attempts used for comparison.


\paragraph{Training and inference:} We fine-tune $\mathcal{M}_0$ on the resulting tutoring contexts and their strategy--feedback targets. At inference, \method{} receives $c_t$ and generates a strategy tag followed by tutor feedback $u_t$ in one call. Only the feedback is delivered to the student, and the tag is retained for analysis. After the student replies, \method{} selects the next strategy using the updated dialogue. AdaTutor thus adapts as the student progresses.

\section{Implementation Details}
\paragraph{Stateful Student Simulator:}
\label{sec:student_sim}

Prompted LLM students can abandon an assigned misconception after feedback that does not address it \citep{dofaithfulness2026}. This makes it difficult to attribute apparent progress to the tutor's assistance. Motivated by prior work advocating explicit representations of student knowledge \citep{yuan2026valid}, we use a stateful LLM-based student simulator. We separate knowledge updates from response generation, drawing on the approach of \citet{pearl2026}. This lets us compare alternative tutor responses from the same dialogue and recorded knowledge state.

The student uses \textit{knowledge component} (KC). It is a reusable knowledge or concept needed for a reasoning step~\citep{corbett1994}. We build a separate KC library for each domain from training problems and reference solutions, drawing on previous work on automated KC generation and tagging \citep{duan2026kc}. For this, we split solutions into reasoning steps and cluster similar steps. We then identify KCs within each cluster and merge overlapping candidates. For selected KCs, we generate incorrect beliefs to construct the \textit{misconception} students \citep{sonkarcognitive2024,dofaithfulness2026}. The library remains fixed during tutoring. We assign each student a simple profile specifying which KCs are mastered, unmastered, or associated with an incorrect belief. \textit{Knowledge-gap} students have unmastered KCs, \textit{misconception} students hold incorrect beliefs about particular KCs, and \textit{mixed} students have both knowledge gaps and misconceptions.

The student's \emph{state} records the current status of the KCs needed for a problem. For example, it may mark addition as mastered but interpreting ''twice as many'' as a misconception, with the belief ''twice means two more.'' We initialize this state from the student's profile. After tutor feedback, two LLM calls run sequentially. The first proposes an updated state based on the feedback. The second generates an answer and a public response conditioned on the state. The state remains fixed during the second call, and only the public response is added to the tutor's dialogue. We check whether responses reflect the assigned state and how they change under different feedback. Appendix~\ref{app:student-simulator} provides more details on KC library construction, misconception generation, and student profiles.

\paragraph{Models and interaction protocol:} We use \texttt{Qwen2.5-32B-Instruct}, and \texttt{Qwen2.5-7B-Instruct} \citep{qwen2025} for student and teacher respectively. All methods begin from the same saved initial student response for each profile and problem, with subsequent responses sampled independently. Each interaction allows at most five tutor turns and ends when the student correctly answers the original problem or reaches the turn limit.

\paragraph{Data collection and training:} We sample $K=10$ independent student attempts to score each candidate response and use $\varepsilon=0.5$ for exploratory continuation. We train a separate \method{} tutor for each dataset using LoRA. We provide all prompts in Appendix~\ref{app:prompts}.

\section{Experiments}
\label{sec:setup}
\paragraph{Datasets and student profiles:} We evaluate on math reasoning on GSM8K \citep{gsm8k2021} and SVAMP \citep{patel-etal-2021-nlp-svamp}. ScienceQA \citep{lu2022learn} contains MCQ's in natural science, social science, and language. We construct 5 student profiles for each of the three student types, amounting to 15 profiles for each dataset.

\paragraph{Baselines and evaluation metrics:} We compare \method{} with six \textit{Fixed-Strategy Tutors} that always use a fixed strategy irrespective of the tutoring context. We also compare with \textsc{Prompt selector}, where the tutor is provided all the six strategies and prompted to select the most appropriate strategy for the context. We also compare with 
SocraticLM~\citep{socraticlm2024}, and TutorRL~\citep{tutorrl2025}. 

We report Cumulative accuracy ($A_t$) to measure the fraction of student--problem pairs solved by turn $t$. We also report Correction rate, defined as the fraction of initially incorrect pairs that are solved by five turns. See Appendix~\ref{app:setup} for additional details on the experimental setup for reproducing the results. Appendix~\ref{app:resp-qeval} evaluates student responses for consistency with assigned knowledge, tests whether student knowledge updates improve performance on new problems, and tutor responses for strategy adherence, correctness, and answer disclosure.


\section{Results and Analysis}
\label{sec:results}

\subsection{Main Tutoring Performance}
\method{} achieves the highest final accuracy among the evaluated tutors, reaching 72.7\%, 82.5\%, and 64.1\% on GSM8K, SVAMP, and ScienceQA after five tutoring turns (Table~\ref{tab:main-results}). \textsc{Worked example} is the strongest fixed strategy on GSM8K and SVAMP, while \textsc{Contrastive feedback} is strongest on ScienceQA. \method{} exceeds these fixed strategies by 6.3, 2.6, and 10.1 percentage points, respectively. It also outperforms the \textsc{Prompt Selector} on all three datasets.

Compared with \textsc{TutorRL}, the gains are 10.4 points on GSM8K, 8.2 on SVAMP, and 3.5 on ScienceQA. The corresponding gains over \textsc{SocraticLM} are 9.8, 13.5, and 7.5 points. These results show that adaptively selecting pedagogical strategies is better than using a fixed strategy, as well as RL-based and Socratic methods. The strongest alternative is \textsc{Worked example} on GSM8K and SVAMP, and \textsc{TutorRL} on ScienceQA. The 95\% confidence intervals for \method{}'s gains over these alternatives are $[4.51,8.05]$, $[0.73,4.48]$, and $[1.63,5.44]$ percentage points, respectively. Paired tests grouped by problem support all three comparisons after Holm correction across the 27 baseline comparisons ($p<0.01$).

\begin{table}[!htb]
\centering
\footnotesize
\setlength{\tabcolsep}{2pt}
\begin{tabularx}{\columnwidth}{@{}>{\raggedright\arraybackslash}X@{\hspace{3pt}}r@{\hspace{3pt}}r@{\hspace{3pt}}r@{}}
\toprule
Tutor & GSM8K & SVAMP & ScienceQA \\
\midrule
Initial accuracy ($A_0$) & 4.3 & 8.5 & 14.4 \\
\midrule
\textsc{Hint} & 8.4 & 16.3 & 41.7 \\
\textsc{Step scaffold} & 32.6 & 32.8 & 51.8 \\
\textsc{Explicit explanation} & 48.9 & 31.3 & 50.0 \\
\textsc{Worked example} & \underline{66.4} & \underline{79.9} & 48.7 \\
\textsc{Self-explanation} & 7.3 & 16.5 & 40.6 \\
\textsc{Contrastive feedback} & 23.0 & 18.8 & \underline{54.0} \\
\midrule
\textsc{TutorRL} {\scriptsize\citep{tutorrl2025}} & 62.3 & 74.3 & 60.6 \\
\textsc{SocraticLM} {\scriptsize\citep{socraticlm2024}} & 62.9 & 69.0 & 56.6 \\
\midrule
\textsc{Prompt Selector} & 18.1 & 34.8 & 50.7 \\
\method{} tutor & \textbf{72.7} & \textbf{82.5} & \textbf{64.1} \\
\midrule
\method{} gain over $A_0$ & +68.3 & +74.0 & +49.7 \\
\method{} correction rate & 71.4 & 80.9 & 58.0 \\
\bottomrule
\end{tabularx}
\caption{Final cumulative accuracy $A_5$ (\%) unless otherwise indicated. The final two rows give \method{}'s gain over initial accuracy in percentage points and its correction rate (\%). Bold marks the highest final accuracy and underlining identifies the strongest fixed strategy. Differences are computed before rounding.}
\label{tab:main-results}
\end{table}


\subsection{Which Strategies does \method{} select?}
The strong fixed-strategy baselines raise a question: does \method{} mainly learn a preferred strategy, assign a strategy to each student type, or vary its choices during the dialogue?

\paragraph{Strategy effectiveness across student types:} Table~\ref{tab:type} compares correction rates for knowledge-gap, misconception, and mixed students. The same fixed strategy is strongest for all three types within each dataset: \textsc{Worked example} on GSM8K and SVAMP, and \textsc{Contrastive feedback} on ScienceQA. \method{} improves on the strongest fixed strategy in eight of the nine comparisons. Its largest gains occur for misconception students: 11.5 points on GSM8K, 6.2 on SVAMP, and 20.4 on ScienceQA. For knowledge-gap students on SVAMP, \textsc{Worked example} reaches 90.9\%, compared with 90.2\% for \method{}.

\begin{table}[!htb]
\centering
\footnotesize
\setlength{\tabcolsep}{2pt}
\renewcommand{\arraystretch}{1.08}
\begin{tabularx}{\columnwidth}{@{}l*{3}{>{\centering\arraybackslash}X}@{}}
\toprule
Dataset & Knowledge gap & Misconception & Mixed \\
\midrule
GSM8K
& $88.3 \rightarrow \mathbf{89.3}$
& $40.5 \rightarrow \mathbf{52.0}$
& $63.7 \rightarrow \mathbf{71.3}$ \\
SVAMP
& $\mathbf{90.9} \rightarrow 90.2$
& $70.4 \rightarrow \mathbf{76.5}$
& $71.1 \rightarrow \mathbf{74.6}$ \\
ScienceQA
& $45.6 \rightarrow \mathbf{53.7}$
& $47.3 \rightarrow \mathbf{67.7}$
& $45.9 \rightarrow \mathbf{54.2}$ \\
\bottomrule
\end{tabularx}
\caption{Correction rates (\%) by student type: Each cell shows the strongest fixed-strategy tutor followed by \method{}, with the higher value in bold. \textsc{Worked example} is strongest across types on GSM8K and SVAMP. \textsc{Contrastive feedback} is strongest across types on ScienceQA. Differences are computed before rounding.}
\label{tab:type}
\end{table}

\begin{figure}[!htb]
\centering
\includegraphics[width=0.65\linewidth]{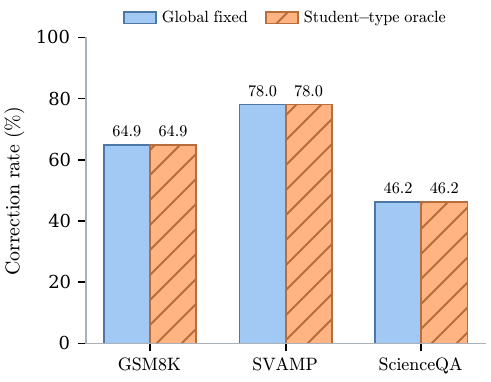}
\caption{Correction rates for the student-type oracle and \method{}. The oracle uses test outcomes to choose a fixed strategy for each student type and matches the globally strongest fixed strategy within each dataset.}
\label{fig:oracle_comparison}
\end{figure}

\paragraph{Applying a fixed strategy per student type:} We next ask whether assigning a different fixed strategy to each student type is enough. We construct a student-type oracle that selects the strongest fixed strategy separately for each type using the test outcomes. The oracle selects the same strategy for every type within each dataset. It therefore matches the globally strongest fixed strategy, with correction rates of 64.9\%, 78.0\%, and 46.2\% on GSM8K, SVAMP, and ScienceQA (Figure~\ref{fig:oracle_comparison}). \method{} reaches 71.4\%, 80.9\%, and 58.0\%, respectively. Assigning a fixed strategy by student type therefore leaves a gap relative to \method{}. We next examine its choices at individual dialogue turns.

\paragraph{Strategy selection across tutoring turns:} On GSM8K, \textsc{Worked example} accounts for 80.5\% of first-turn choices and 35.7\% at turn five. By turn five, \textsc{Contrastive feedback}, \textsc{Hint}, and \textsc{Step scaffold} account for 32.9\%, 17.0\%, and 13.7\%, respectively. On SVAMP, \textsc{Worked example} remains the most frequent strategy, decreasing from 82.4\% to 62.2\%, while \textsc{Contrastive feedback} increases from 6.8\% to 30.4\%. ScienceQA shows a different pattern. \textsc{Step scaffold} and \textsc{Hint} account for 32.7\% and 27.1\% of first-turn choices. \textsc{Hint} becomes the most frequent strategy at turn five, at 40.4\%. The changing mix of demonstrations, corrections, and hints illustrates the choice among forms of assistance raised by the \textit{assistance dilemma}. Choices also differ across interactions at the same turn number. See Appendix for Figures~\ref{fig:selector-tags-gsm8k} (GSM8K), ~\ref{fig:selector-tags-svamp}(SVAMP) and~\ref{fig:selector-tags-scienceqa}(ScienceQA).

\subsection{Does Context Help Strategy Selection?}
We first examine how consecutive strategy choices are related. We then test whether access to the dialogue helps the tutor select strategies and generate useful feedback.

\paragraph{Consecutive strategy choices:} Across datasets, adjacent tutor turns use different strategies in 54.3--62.3\% of comparisons. Among sessions reaching at least two tutor turns, 67.3--84.9\% include a strategy change (Table~\ref{tab:strategy_switching}). Conditional mutual information between consecutive choices ranges from 0.08 to 0.19 bits after accounting for student type and turn. This describes an association between successive choices beyond those two variables.

\begin{table}[!htb]
\centering
\small
\setlength{\tabcolsep}{2.5pt}
\renewcommand{\arraystretch}{1.05}
\begin{tabularx}{\columnwidth}{
@{}
>{\raggedright\arraybackslash}X
>{\centering\arraybackslash}p{0.30\columnwidth}
>{\centering\arraybackslash}p{0.25\columnwidth}
>{\centering\arraybackslash}p{0.27\columnwidth}
@{}}
\toprule
Dataset & Switch Rate (\%) & Sessions (\%) & Cond. MI (bits) \\
\midrule
GSM8K     & 62.3 & 79.2 & 0.17 \\
SVAMP     & 54.3 & 67.3 & 0.08 \\
ScienceQA & 56.0 & 84.9 & 0.19 \\
\bottomrule
\end{tabularx}
\caption{Strategy changes during tutoring. Switch rate uses adjacent tutor-turn pairs. Sessions with a switch uses sessions that reach at least two tutor turns. Conditional mutual information measures the association between consecutive strategies after accounting for student type and turn.}
\label{tab:strategy_switching}
\end{table}

\paragraph{Contribution of dialogue context:} We now test whether dialogue context also adds to tutoring performance. For each dataset, we train four tutors that differ in the dialogue they receive: the full history, the latest tutor--student exchange, the latest student reply, or no dialogue. All four receive the problem and reference solution. Each restriction applies during both training and inference.

\begin{table}[!htb]
\centering
\small
\setlength{\tabcolsep}{3pt}
\begin{tabularx}{\columnwidth}{@{}>{\raggedright\arraybackslash}Xrrr@{}}
\toprule
Context & GSM8K & SVAMP & ScienceQA \\
\midrule
Full history & 72.7 & \textbf{80.2} & 63.9 \\
Latest exchange & \textbf{72.8} & 78.1 & 65.4 \\
Latest student reply & 72.4 & 79.2 & 65.2 \\
Problem only & 50.6 & 72.3 & \textbf{66.7} \\
\bottomrule
\end{tabularx}
\caption{Final accuracy $A_5$ (\%) with different amounts of dialogue context. All variants receive the problem and reference solution and are trained separately under the same checkpoint-selection rule. Bold marks the highest accuracy per dataset. Differences are computed before rounding.}
\label{tab:context}
\end{table}

Dialogue improves tutoring on both arithmetic datasets (Table~\ref{tab:context}). On GSM8K, removing it lowers accuracy from 72.7\% to 50.6\%, a decrease of 22.2 points. Using only the latest exchange reaches 72.8\%, close to the full-history result. On SVAMP, full history reaches 80.2\%, compared with 72.3\% for the problem-only tutor, a difference of 8.0 points. ScienceQA instead favors the problem-only tutor, which reaches 66.7\% against 63.9\% with full history. The best amount of dialogue context therefore differs across datasets.

\subsection{What Does \method{} Learn?}
\method{} learns both which strategy to select and what feedback to provide. To examine their contributions, we compare the first tutor turn, when the methods receive the same initial student reply. For each initially incorrect case, we take the strategy selected by \method{} or the \textsc{Prompt Selector} and use the corresponding fixed-strategy tutor's feedback. Both methods then use the same feedback for a given strategy and context, allowing us to compare their choices. We also evaluate each method's own generated feedback.

With fixed-strategy feedback, \method{} improves first-turn correction over the \textsc{Prompt Selector} by 34.1, 33.8, and 13.8 points on GSM8K, SVAMP, and ScienceQA (Table~\ref{tab:selection-realization}). Using \method{}'s own feedback adds 2.1, 3.5, and 7.9 points, respectively. In these first-turn comparisons, the difference associated with strategy selection is larger, while response generation provides a further improvement.

\begin{table}[!htb]
\centering
\small
\setlength{\tabcolsep}{3pt}
\renewcommand{\arraystretch}{1.03}
\begin{tabularx}{\columnwidth}{@{}l>{\raggedright\arraybackslash}Xcc@{}}
\toprule
Dataset & Tutor & Fixed (\%) & Own (\%) \\
\midrule
\multirow{2}{*}{GSM8K}
& \textsc{Prompt Selector} & 2.9 & 4.7 \\
& \method{} & \textbf{37.0} & \textbf{39.1} \\
\midrule
\multirow{2}{*}{SVAMP}
& \textsc{Prompt Selector} & 9.2 & 13.8 \\
& \method{} & \textbf{43.0} & \textbf{46.6} \\
\midrule
\multirow{2}{*}{ScienceQA}
& \textsc{Prompt Selector} & 21.9 & 27.7 \\
& \method{} & \textbf{35.6} & \textbf{43.5} \\
\bottomrule
\end{tabularx}
\caption{First-turn correction rates among initially incorrect cases. Fixed response uses feedback from the fixed-strategy tutor corresponding to the selected strategy. Own response uses each method's generated feedback. Differences are computed before rounding.}
\label{tab:selection-realization}
\end{table}

\subsection{Qualitative Examples}
\begin{figure*}[!htb]
\centering
\includegraphics[width=\textwidth]{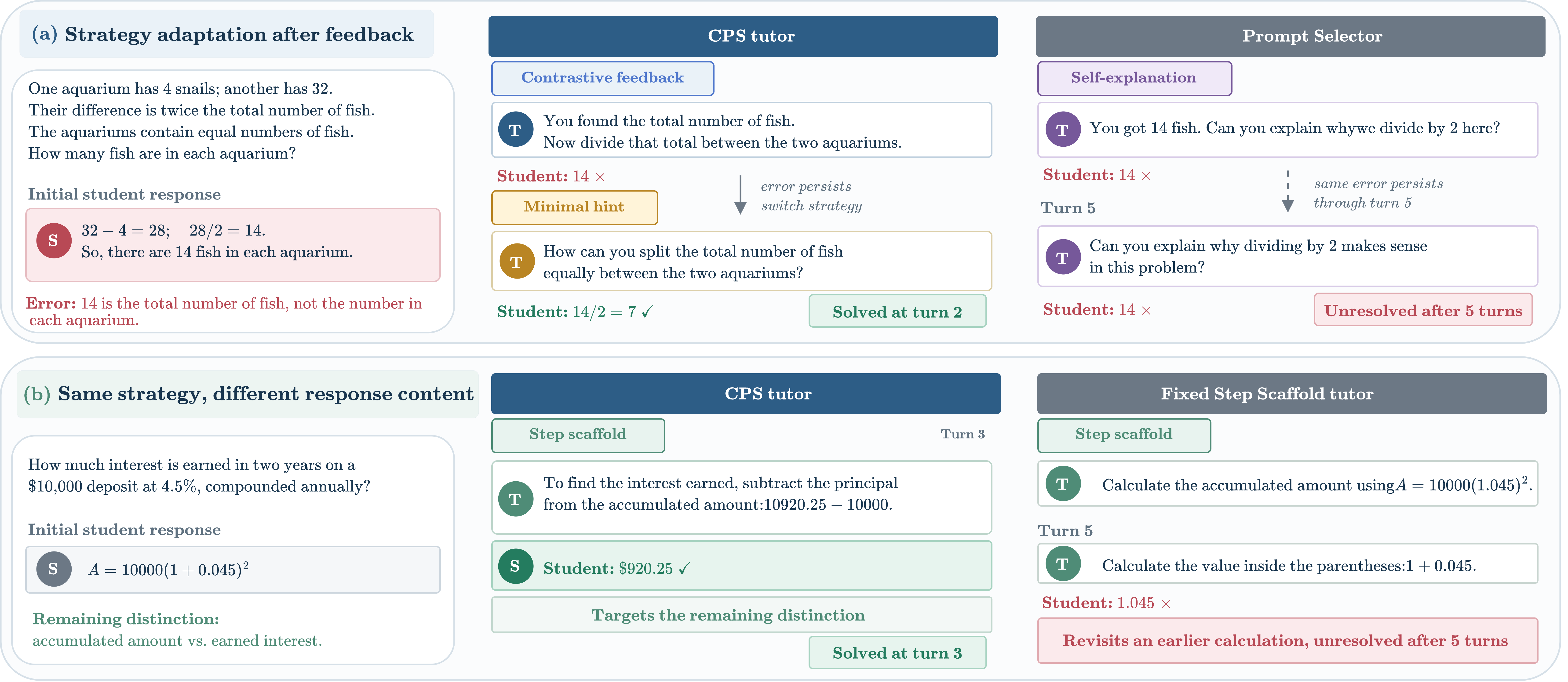}
\caption{\textbf{Qualitative tutoring examples:} \textbf{(a)} \method{} changes from \textsc{Contrastive feedback} to \textsc{Hint} after the initial intervention does not resolve the student's error, while the \textsc{Prompt Selector} continues eliciting explanation of the same incorrect step. \textbf{(b)} Both tutors use \textsc{Step scaffold}, but \method{} targets the student's remaining confusion between accumulated amount and earned interest, while the fixed-strategy tutor continues scaffolding an earlier calculation. Strategy labels are retained for analysis but are not shown to the student.}
\label{fig:qualitative}
\end{figure*}

Figure~\ref{fig:qualitative} illustrates two complementary behaviors. In (a), \method{} changes its pedagogical strategy after the initial feedback does not resolve the student's error. In (b), the comparison shows that the effectiveness of a strategy also depends on how it is realized. Both tutors use \textsc{Step scaffold}, but \method{} targets the student's remaining error while the fixed-strategy tutor continues scaffolding an earlier calculation.

\subsection{Transfer Across Datasets and Tutors}
The previous analyses examine tutoring within each dataset. We now test whether the learned tutoring behavior transfers to another dataset and whether the strategy choices can guide a different tutor model.

\paragraph{Cross-dataset transfer:} We evaluate each trained \method{} tutor on the other two datasets without further training or checkpoint selection on the target dataset. For each target, all tutors use the same student profiles, problems, initial replies, and five-turn budget. Table~\ref{tab:ood} reports final accuracy.

Transfer is strongest between GSM8K and SVAMP. The SVAMP-trained tutor reaches 75.5\% on GSM8K, compared with 72.7\% for the GSM8K-trained tutor and 66.4\% for the strongest fixed strategy. In the reverse direction, the GSM8K-trained tutor reaches 78.7\% on SVAMP, 3.8 points below the in-domain tutor and 1.2 points below the strongest fixed strategy. These arithmetic datasets also share the same KC library and student profiles.

Transfer between arithmetic and ScienceQA yields lower accuracy. Tutors trained on GSM8K and SVAMP reach 49.1\% and 51.1\% on ScienceQA, compared with 64.1\% for the ScienceQA-trained tutor. The ScienceQA-trained tutor reaches 33.5\% on GSM8K and 31.1\% on SVAMP. This difference is expected given the differences between the domains. Arithmetic and science involve different reasoning patterns, student difficulties, and misconceptions, which may require different pedagogical strategies. The stronger transfer between GSM8K and SVAMP suggests that the learned tutoring behavior transfers best when the source and target domains share similar student needs.


\begin{table}[!htb]
\centering
\small
\setlength{\tabcolsep}{3pt}
\begin{tabularx}{\columnwidth}{@{}>{\raggedright\arraybackslash}Xrrr@{}}
\toprule
Source dataset & GSM8K & SVAMP & ScienceQA \\
\midrule
GSM8K & \underline{72.7} & 78.7 & 49.1 \\
SVAMP & \textbf{75.5} & \underline{\textbf{82.5}} & 51.1 \\
ScienceQA & 33.5 & 31.1 & \underline{\textbf{64.1}} \\
\midrule
\makecell[l]{Strongest fixed\\strategy} & 66.4 & 79.9 & 54.0 \\
\bottomrule
\end{tabularx}
\caption{Cross-dataset final accuracy $A_5$ (\%). Rows identify the tutor's training dataset. Columns identify the evaluation dataset. Underlining denotes in-domain evaluation, and bold marks the highest trained-tutor accuracy per target. The target's strongest fixed strategy is shown for comparison.}
\label{tab:ood}
\end{table}

\paragraph{Transfer to an API tutor:} To test whether learned strategy choices can guide a different response generator, we pair a GSM8K-trained \method{} tutor with GPT-4.1-mini. At each turn, we use only \method{}'s strategy tag and ask GPT-4.1-mini to generate the feedback. The comparison includes six fixed-strategy tutors and the \textsc{Prompt Selector}, all using GPT-4.1-mini to generate feedback. To save cost, we use 247 GSM8K examples for this experiment.

With \method{}'s strategy choices, GPT-4.1-mini reaches 89.9\% final accuracy, at least 7.3 points above every fixed-strategy tutor (Table~\ref{tab:api}). Relative to the API \textsc{Prompt Selector}, the gain is 5.3 points after one turn, with a paired 95\% confidence interval of $[1.2,9.7]$. The final gain is 2.0 points, with an interval of $[-2.0,6.1]$. The clearest advantage is therefore over fixed strategies and in the first-turn comparison with prompted selection.

The benefit also varies by student type. For misconception students, \method{}'s strategy choices yield 91.6\% final accuracy, compared with 84.2\% for the API \textsc{Prompt Selector}. The methods tie at 86.4\% for knowledge-gap students. For mixed students, they reach 92.2\% and 95.3\%, respectively.

\begin{table}[!htb]
\centering
\small
\setlength{\tabcolsep}{3pt}
\begin{tabularx}{\columnwidth}{@{}>{\raggedright\arraybackslash}Xrrr@{}}
\toprule
Tutor & $A_1$ & $A_5$ & \makecell{Correction (\%)} \\
\midrule
\textsc{Step scaffold} & 64.0 & 82.6 & 58.3 \\
\textsc{Contrastive feedback} & 72.9 & 82.6 & 58.3 \\
\textsc{Prompt Selector} & 68.0 & 87.9 & 70.9 \\
\method{} selection + API & \textbf{73.3} & \textbf{89.9} & \textbf{75.7} \\
\bottomrule
\end{tabularx}
\caption{Strategy selection with GPT-4.1-mini as the response generator (\%). This earlier GSM8K study uses 247 assignments with $A_0=58.3\%$, rather than the current main evaluation. The two strongest fixed strategies are shown. Differences are computed before rounding.}
\label{tab:api}
\end{table}

\subsection{Response Quality Evaluation}
We use an LLM Judge to evaluate 270 initial student replies, 180 replies during tutoring, and 180 fixed-strategy tutor responses on GSM8K. It judges 95.0\% of tutor responses mathematically correct and identifies no correct final-answer disclosure. On examples also rated by humans, the agreement is 84.0\% for student-KC judgments and 95.0\% for tutor strategy and correctness judgments. We also test whether student-state updates help on new problems. Using the state after relevant feedback raises accuracy on 300 new GSM8K problems that need the same KCs, from 5.0\% to 67.7\%, with the teaching dialogue removed and model parameters fixed. See Appendix~\ref{app:resp-qeval} for more details.

\section{Conclusion}
We introduced \method{}, a framework to train a tutor to adapt its pedagogical strategy to the student's work and the preceding dialogue. We train it by comparing feedback that uses different strategies at the same point in a tutoring session. The selected strategy and feedback provide supervision for a tutor that chooses a strategy and generates its response in one call. Experiments on GSM8K, SVAMP, and ScienceQA show significant improvements over the strongest alternative tutor after five turns. First-turn comparisons show contributions from both learned strategy choices and feedback generation. Its learned strategy choices also improve tutoring with GPT-4.1-mini over fixed strategies. These findings support adapting the form of assistance to the current tutoring context rather than retaining one strategy throughout a session.

\section*{Limitations}
Our experiments provide an efficient testbed for comparing pedagogical strategies with simulated students. Current results do not demonstrate human learning.


\section*{Acknowledgments}

\bibliography{custom}

\appendix

\section{Pedagogical Strategies}
\label{app:strategies}

Table~\ref{tab:strategies} defines six pedagogical strategies and illustrates each with an example problem and an initial student response. Each strategy identifies the primary form of assistance, with the specific content varying by tutoring context. These strategies serve as alternatives, not as a prescribed sequence or a fixed mapping from student types to instructional responses.

\begin{table*}[!htb]
\centering
\small
\setlength{\tabcolsep}{5pt}
\renewcommand{\arraystretch}{1.12}
\begin{tabularx}{\textwidth}{@{}>{\raggedright\arraybackslash}p{0.16\textwidth}>{\raggedright\arraybackslash}p{0.31\textwidth}>{\raggedright\arraybackslash}X@{}}
\toprule
\multicolumn{3}{@{}p{\textwidth}@{}}{\textbf{Problem:} Mia has 6 marbles. Leo has twice as many. Each person receives 3 more marbles. How many marbles do they have altogether?} \\
\addlinespace[4pt]
\multicolumn{3}{@{}p{\textwidth}@{}}{\textbf{Initial student response:} "Leo has $6+2=8$ marbles because 'twice as many' means two more.''} \\
\midrule
\textbf{Strategy} & \textbf{Definition} & \textbf{Illustrative tutor response} \\
\midrule

\textsc{Minimal hint} & Provide a brief cue directing attention to the relevant concept or relationship, allowing the student to independently determine the subsequent step. & "Focus on the word 'twice' when comparing Leo's marbles with Mia's." \\
\addlinespace[6pt]

\textsc{Step scaffold} & Guide the next step by breaking it into smaller actions or providing a partial setup. & "First find Leo's starting number of marbles. This is two groups of 6. What is $2\times6$?" \\
\addlinespace[6pt]

\textsc{Explicit explanation} & Explain the relevant concept or rule and the reasoning behind it. & "'Twice as many' means two equal groups of the original amount. Multiplication by two expresses this relationship." \\
\addlinespace[6pt]

\textsc{Worked example} & Demonstrate the relevant reasoning on an analogous problem with different values. Then, ask the student to apply the same method. & "Suppose Ana has 4 marbles and Ben has twice as many. Ben has $2\times4=8$. If each receives 1 more, their total is $(4+1)+(8+1)=14$. Apply the same steps to Mia and Leo." \\
\addlinespace[6pt]

\textsc{Self-explanation} & Ask the student to explain or justify their reasoning, making their interpretation explicit. & "You used $6+2$. What does adding 2 represent in your reasoning, and why did you choose it for 'twice as many'?" \\
\addlinespace[6pt]

\textsc{Contrastive feedback} & Compare the student's incorrect reasoning with a correct alternative and explain the distinction. & "Your $6+2$ represents two extra marbles. 'Twice as many' instead means two groups of 6, represented as $2\times6$. The distinction is adding two objects versus doubling the original group." \\

\bottomrule
\end{tabularx}
\caption{Definitions and examples of the six pedagogical strategies. Each row illustrates an example response of the respective strategy to the same tutoring context. All strategies withhold the final answer to the original problem.}
\label{tab:strategies}
\end{table*}

\section{Stateful Student Simulator}
\label{app:student-simulator}

The simulator uses a KC library, a small library of screened misconceptions, synthetic student profiles, and a two-call LLM interaction. These components create controlled student conditions for tutor training and evaluation rather than attempting to model the full behavior of human students.

\begin{figure*}[!htb]
\centering
\includegraphics[width=0.8\textwidth]{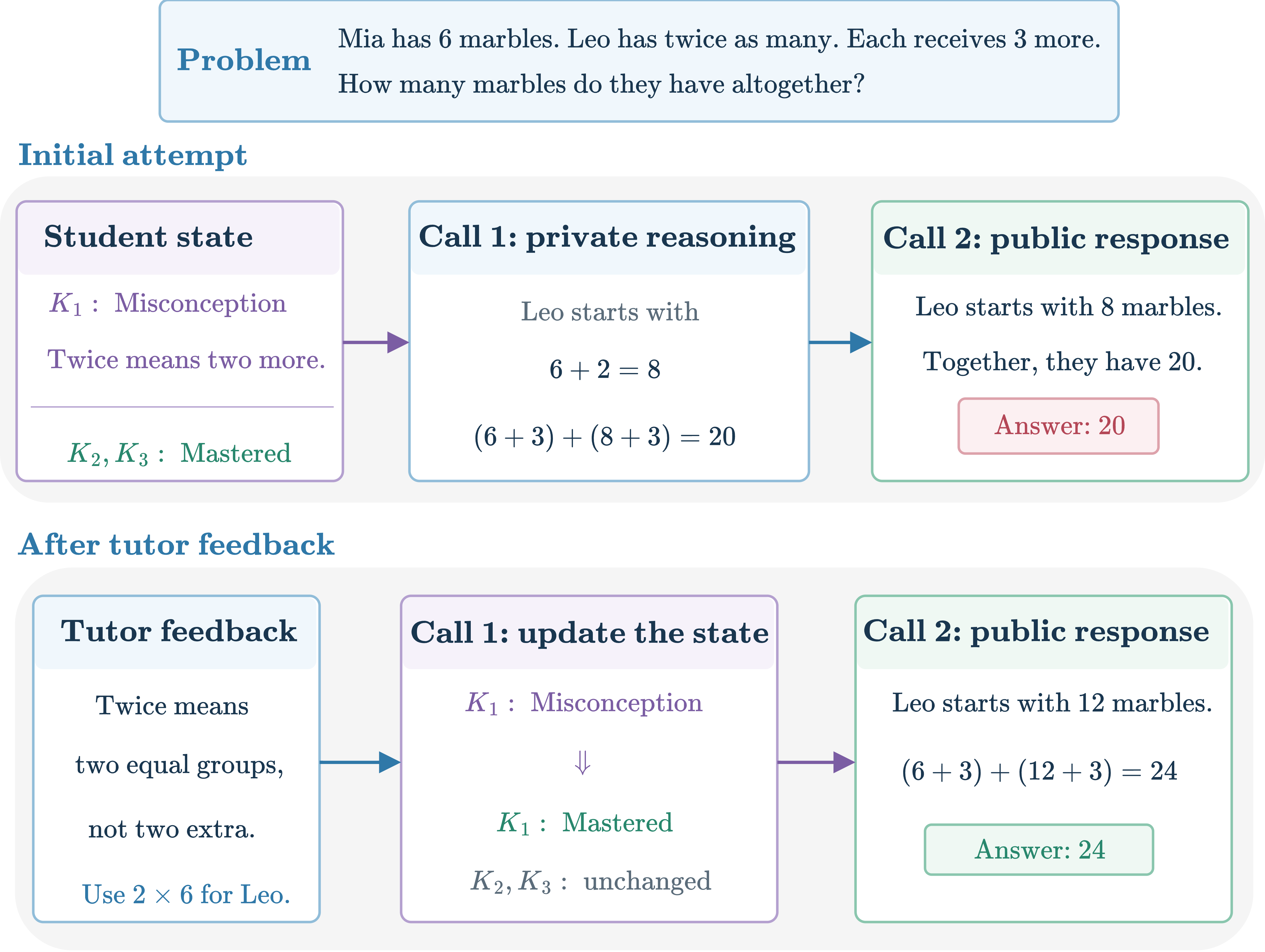}
\caption{\textbf{Student simulation}: The initial attempt uses one call for private reasoning and another for the public reply. After tutor feedback, one call updates the KC state, and another generates the reply and answer under that state. Only the public reply enters the tutor's dialogue. The example illustrates the procedure rather than reporting an experimental result.}
\label{app:fig:student-simulator}
\end{figure*}

\begin{figure*}[!htb]
\centering
\includegraphics[width=0.8\textwidth]{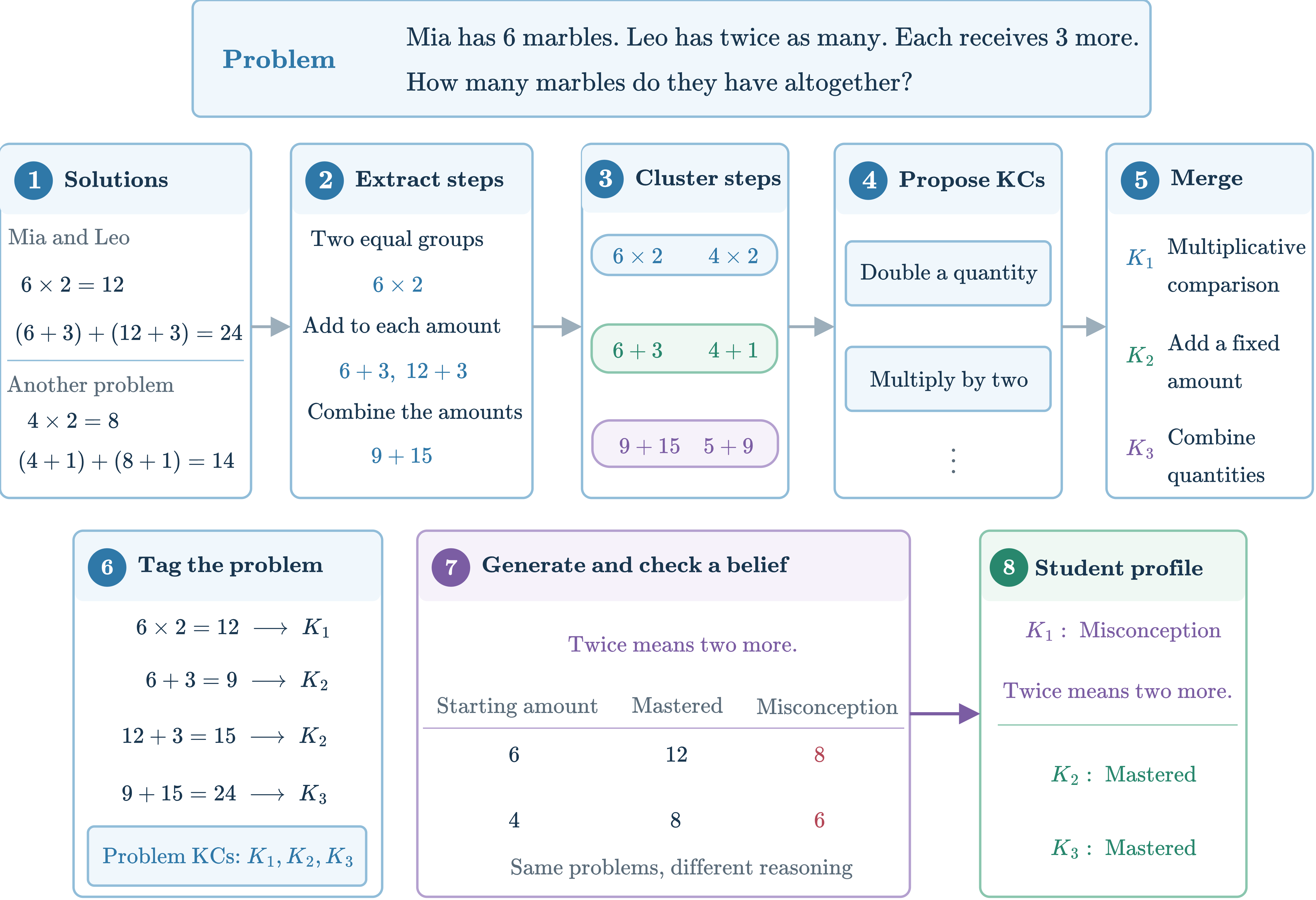}
\caption{\textbf{Dataset-specific KC construction:} We group similar reasoning steps from training solutions, identify reusable KCs, and merge overlapping definitions. The resulting library is used to annotate problems. Incorrect beliefs are checked on problems requiring the corresponding KC before they are included in student profiles. The figure uses an illustrative example.}
\label{app:fig:kc-construction}
\end{figure*}

\paragraph{KC library and misconception construction:} Knowledge components have long been used to represent fine-grained student knowledge \citep{corbett1994}. We construct them automatically from training solutions, following the general LLM-based KC generation and problem-tagging approach of \citet{duan2026kc}. We decompose each reference solution into small reasoning steps. We then cluster steps with similar meaning, propose local KC candidates, and merge synonymous or redundant candidates. We retain recurring KCs and review the resulting library before using it for problem annotation. We then annotate each problem with the KCs it needs using the reference solution. Figure~\ref{app:fig:kc-construction} summarizes this procedure.

Not every KC used for problem annotation is used to define students. We retain KCs that occur often enough and are suitable for controlling student knowledge. For these KCs, an LLM proposes several reusable incorrect beliefs. We remove malformed, duplicate, or overly specific candidates and test the remaining beliefs on problems requiring the corresponding KC. Following prior work on simulated students with misconceptions \citep{sonkarcognitive2024} and concerns about misconception faithfulness \citep{dofaithfulness2026}, we retain a misconception only when the student actually follows the intended incorrect belief and its performance decreases relative to a mastered control on the same problems.

\paragraph{Student profiles:} A student profile assigns one of three states to the KCs used in the experiment: \textit{mastered}, \textit{unmastered}, or \textit{misconception}. A \textit{knowledge-gap} student has one or more unmastered KCs, a \textit{misconception} student has one or more assigned incorrect beliefs, and a \textit{mixed} student has both. We use a fully mastered profile when screening gaps and misconceptions. A profile can contain several deficits, but for a particular problem the student receives only the states of the KCs needed by that problem. State changes persist during that tutoring interaction, and the profile is used again to initialize the next problem. Dataset-specific KCs and profile counts are reported in Appendix~\ref{app:setup}.

\paragraph{Student responses:} The student uses two sequential LLM calls, similar in spirit to prior work that separates student state from response generation \citep{pearl2026}. For the initial attempt, the first call produces private reasoning and an answer under the assigned KC states. The second call turns that fixed result into the public student response. After tutor feedback, the first call decides whether the feedback changes any relevant KC state, and the second attempts the original problem under the resulting state. The runtime checks required fields and state labels but does not execute equations, recover answers from the public response, or repair the model's mathematics. Only the public response reaches the tutor. The answer to the original problem is checked separately for correctness. Figure~\ref{app:fig:student-simulator} shows the interaction and an illustrative state update.

\paragraph{Simulator checks:} LLM students can revert to their underlying competence instead of following assigned gaps or misconceptions \citep{yuan2026valid,dofaithfulness2026}. Recent work therefore evaluates student simulators beyond surface response quality \citep{scarlatossimulation2026}. We separately check whether responses follow the active KC states, whether the public response agrees with the returned answer, and whether actual state changes are supported by tutor feedback. Appendix~\ref{app:student-resp-qeval} reports the detailed checks and results. These checks support the simulator as a controlled synthetic environment.

\section{Experimental Setup}
\label{app:setup}

\paragraph{Datasets and student profiles:} During collection, each problem is paired with one profile containing a relevant knowledge gap or misconception. For evaluation, we use five knowledge-gap, five misconception, and five mixed profiles. GSM8K and SVAMP share the same KC library and profiles. Their knowledge-gap and misconception profiles each contain four affected KCs, while mixed profiles contain two gaps and two misconceptions. We evaluate each profile on 200 distinct problems involving at least one of its assigned gaps or misconceptions. The selected problems can differ across profiles. Table~\ref{app:tab:setup},~\ref{app:tab:training-records} reports the student and dataset statistics.

\begin{table}[!htb]
\centering
\small
\setlength{\tabcolsep}{3pt}
\begin{tabularx}{\columnwidth}{@{}>{\raggedright\arraybackslash}Xrrr@{}}
\toprule
 & GSM8K & SVAMP & ScienceQA \\
\midrule
KCs in annotation library & 34 & 34 & \textit{TBC} \\
KCs eligible for profile construction & 28 & 28 & \textit{TBC} \\
Affected KCs per profile & 4 & 4 & \textit{TBC} \\
\midrule
Knowledge-gap profiles & 5 & 5 & 5 \\
Misconception profiles & 5 & 5 & 5 \\
Mixed profiles & 5 & 5 & 5 \\
Problems per profile & 200 & 200 & 200 \\
Test assignments & 3,000 & 3,000 & 3,000 \\
Unique test problems & 1,289 & 853 & 749 \\
\bottomrule
\end{tabularx}
\caption{Current test populations. An assignment pairs a student profile with an applicable problem. The same problem can be evaluated under more than one profile. Mastered controls are used for screening rather than the main tutoring comparison.}
\label{app:tab:setup}
\end{table}

\begin{table}[!htb]
\centering
\small
\setlength{\tabcolsep}{3.5pt}
\begin{tabular}{@{}lrrr@{}}
\toprule
Setting & GSM8K & SVAMP & ScienceQA \\
\midrule
Training assignments & 2,934 & 3,000 & 8,624 \\
Validation assignments & 485 & 375 & 1,606 \\
Training SFT examples & 2,760 & 2,998 & 6,112 \\
Validation SFT examples & 469 & 378 & 1,111 \\
Training epochs & 3 & 4 & 4 \\
Maximum sequence length & 4,096 & 8,192 & 4,096 \\
Selected checkpoint step & 519 & 720 & 750 \\
Validation $A_5$ (\%) & 76.1 & 83.2 & 66.7 \\
\bottomrule
\end{tabular}
\caption{Training records for the current GSM8K, SVAMP, and ScienceQA tutors. Sequence lengths are in tokens. Assignment counts and SFT-example counts measure different units.}
\label{app:tab:training-records}
\end{table}

\paragraph{Models and Interaction Protocol:} Table~\ref{app:tab:decoding} lists the decoding temperatures. In the two-call simulator, state updating and subsequent response generation use separate decoding settings.

\begin{table}[!htb]
\centering
\small
\setlength{\tabcolsep}{5pt}
\begin{tabular}{@{}lr@{}}
\toprule
Generation stage & Temperature \\
\midrule
Initial student attempt & 0.0 \\
Tutor response & 0.3 \\
Student state update & 0.0 \\
Subsequent student response & 0.3 \\
Candidate-scoring student attempt & 0.3 \\
\bottomrule
\end{tabular}
\caption{Decoding temperatures by generation stage. Student state updates and public response generation are separate calls.}
\label{app:tab:decoding}
\end{table}

\paragraph{Data Collection and Training:} The collection and training configuration is selected through GSM8K validation experiments, and each dataset's checkpoint is selected using final accuracy in five-turn validation interactions. Table~\ref{app:tab:training} lists the training settings. Appendix~\ref{app:recipe} reports the collection and training ablations.

\begin{table}[!htb]
\centering
\small
\setlength{\tabcolsep}{5pt}
\begin{tabular}{@{}lr@{}}
\toprule
Setting & Value \\
\midrule
LoRA rank & 16 \\
LoRA scaling factor & 32 \\
LoRA dropout & 0.05 \\
Learning rate & $2\times10^{-4}$ \\
Learning-rate schedule & Cosine decay \\
Warmup & 5\% \\
Maximum sequence length & 4,096 tokens \\
Training precision & bfloat16 \\
\bottomrule
\end{tabular}
\caption{\method{} fine-tuning configuration.}
\label{app:tab:training}
\end{table}

\paragraph{Baselines and evaluation metrics:} Each fixed-strategy tutor adapts its response content while retaining one strategy throughout the interaction. The \textsc{Prompt selector} chooses among the same six strategies and generates a response at each turn. We use the released 7B checkpoints for TutorRL and SocraticLM with their native prompt formats. In the GSM8K comparison, TutorRL receives the problem and public dialogue. SocraticLM additionally receives the reference answer and solution analysis. The fixed-strategy tutors, \textsc{Prompt Selector}, and \method{} receive the problem, reference solution, and public dialogue. We report accuracy at each turn, final accuracy $A_5$, gain over initial accuracy $A_5-A_0$, and the correction rate. Analyses by student type use only profiles of that type. These metrics measure assisted solving of the original problem. Accuracies and correction rates are reported as percentages, and gains as percentage points.

Let $\mathcal{E}$ denote the evaluated student--problem pairs. Let $Y_{i,p,t}^{(m)}$ equal one if profile $i$ has solved problem $p$ by or before tutor turn $t$ under method $m$, and zero otherwise. Turn zero denotes the initial attempt. Cumulative accuracy is
\begin{equation}
A_t^{(m)}
=
\frac{1}{|\mathcal{E}|}
\sum_{(i,p)\in\mathcal{E}}Y_{i,p,t}^{(m)}.
\label{eq:accuracy}
\end{equation}
Each profile contributes 200 assignments. For student-type analyses, we restrict $\mathcal{E}$ to profiles of that type. The correction rate, computed on the same evaluation population, is
\begin{equation}
    \mathrm{CorrectionRate}^{(m)}
    =
    \frac{A_5^{(m)}-A_0}{1-A_0}
    \label{eq:correction}
\end{equation}
The equations use proportions before conversion to the percentages reported in the tables.

\section{Results \& Analysis}

\begin{figure}[!htb]
\centering
\includegraphics[width=\linewidth]{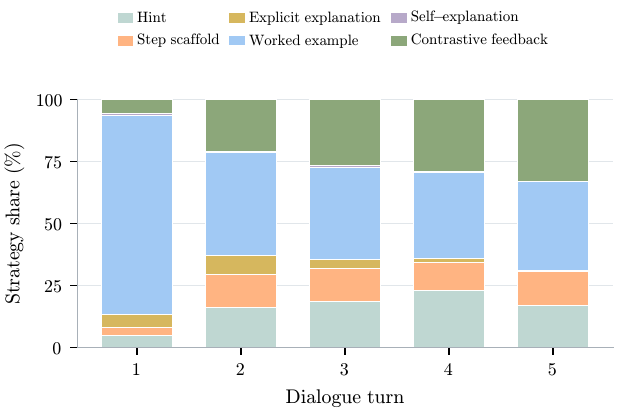}
\caption{Pedagogical strategy choices by tutoring turn on GSM8K. Percentages use valid strategy tags from interactions still active at each turn.}
\label{fig:selector-tags-gsm8k}
\end{figure}

\begin{figure}[!htb]
\centering
\includegraphics[width=\linewidth]{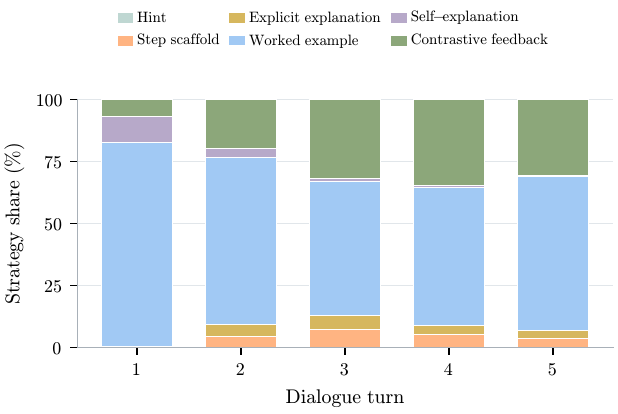}
\caption{Pedagogical strategy distributions by tutoring turn on SVAMP. Percentages are computed over valid strategy tags from interactions still active at each turn.}
\label{fig:selector-tags-svamp}
\end{figure}

\begin{figure}[!htb]
\centering
\includegraphics[width=\linewidth]{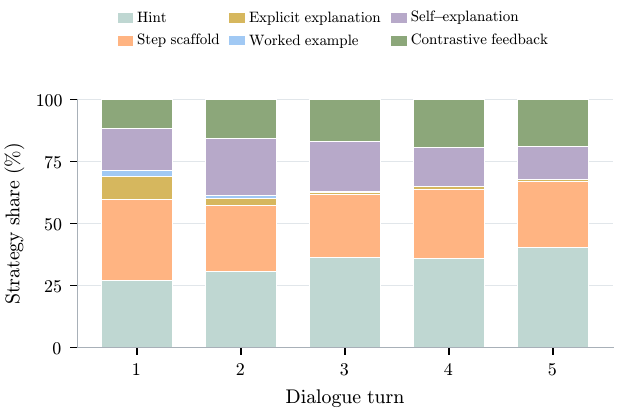}
\caption{Pedagogical strategy distributions by tutoring turn on ScienceQA. Percentages are computed over valid strategy tags from interactions still active at each turn.}
\label{fig:selector-tags-scienceqa}
\end{figure}

\section{Collection and Training Ablations}
\label{app:recipe}

We investigate how different collection and training choices affect the performance of the \method{} tutor. All experiments use GSM8K validation interactions with the same 297 problem assignments and shared initial student responses ($A_0=65.32\%$). We evaluate each configuration through five-turn tutoring interactions.

The experiments are conducted sequentially. Each stage examines one design choice while retaining the configuration selected in the preceding stages. Table~\ref{tab:recipe} summarizes the results.

\begin{table*}[!htb]
\centering
\small
\setlength{\tabcolsep}{7pt}
\renewcommand{\arraystretch}{1.05}
\begin{tabular}{lrr}
\toprule
Configuration & SFT examples & Validation $A_5$ \\
\midrule

\multicolumn{3}{l}{\textit{Outcome construction}} \\
Direct completion, $K=10$
    & 773 & \textbf{88.55} \\
One student response followed by ten probes
    & 586 & 70.37 \\
Ten student responses with paired probes
    & 667 & 71.72 \\

\midrule
\multicolumn{3}{l}{\textit{Trajectory continuation}} \\
$\varepsilon=0.0$
    & 773 & 88.55 \\
$\varepsilon=0.1$
    & 798 & 87.21 \\
$\varepsilon=0.2$
    & 904 & 88.55 \\
$\varepsilon=0.5$
    & 1,048 & \textbf{90.24} \\

\midrule
\multicolumn{3}{l}{\textit{Supervised targets and sampling}} \\
Unique positive winners
    & 1,048 & \textbf{90.24} \\
Include one seeded positive tie
    & 1,399 & 86.20 \\
Winner margin $\geq 0.2$
    & 869 & 89.56 \\
Winner margin $\geq 0.5$
    & 567 & 86.53 \\
Strategy-balanced sampling
    & 1,048 & 79.80 \\
Turn--strategy-balanced sampling
    & 1,048 & 75.42 \\

\bottomrule
\end{tabular}
\caption{
Sequential training ablations on GSM8K validation interactions. All configurations are evaluated on the same initial student responses. Results use checkpoints selected by minimum validation loss. Bold indicates the selected configuration within each comparison group.
}
\label{tab:recipe}
\end{table*}

\subsection{Outcome Construction}

We compare three approaches for estimating the outcome of each strategy-conditioned response. The first directly evaluates whether the student can solve the original problem after receiving the candidate response, using ten independently sampled attempts.

The other two approaches introduce an intermediate student response before evaluating completion. The first samples one student response followed by ten completion probes. The second samples ten student responses, scoring each directly when it answers the problem or using a paired completion probe otherwise.

Direct completion produces unique positive winners in 30.24\% of collection histories, compared with 18.00\% and 20.87\% for the two alternative approaches. It also produces substantially fewer tied maxima: 8.96\%, compared with 68.94\% and 66.43\%.

The resulting supervision yields considerably higher downstream tutoring accuracy. With identical initial student responses, direct completion reaches 88.55\% validation accuracy, compared with 70.37\% and 71.72\% for the alternatives.

Direct completion also reduces collection cost. It requires 191,391 model requests, compared with 239,414 and 432,345 for the other approaches. These findings motivate using direct completion value as the supervision signal in \method{}.

\subsection{Exploratory Trajectory Continuation}
We next examine the effect of exploration during data collection. The supervised target at each tutoring context remains the highest-scoring candidate. The continuation policy determines which candidate response advances the public dialogue and consequently which tutoring contexts are encountered at later turns.

We compare four values of $\varepsilon$, ranging from greedy continuation ($\varepsilon=0$) to exploration with
$\varepsilon=0.5$.

Increasing exploration exposes the tutor to additional histories following responses that do not achieve the highest observed direct completion value. At $\varepsilon=0.5$, the collection produces 1,048 supervised examples, compared with 773 under greedy continuation.

The resulting tutor achieves 90.24\% validation accuracy, the highest observed result among the evaluated configurations. This represents an improvement of 1.68 percentage points over greedy continuation.

We adopt $\varepsilon=0.5$ based on its validation performance and increased coverage of subsequent tutoring contexts.

\subsection{Supervised Targets and Sampling}
We investigate whether modifying the supervised target selection criterion or training-data sampling improves
the distilled tutor.

Our default approach retains comparisons with a unique positive maximum. We compare it with three alternatives: including one seeded candidate from positive ties and requiring winner margins of at least 0.2 or 0.5.

Including positive ties increases the training set from 1,048 to 1,399 examples but reduces validation accuracy from 90.24\% to 86.20\%. Requiring larger winner margins also reduces the number of available supervised examples. A margin of 0.2 produces 89.56\% validation accuracy, while a margin of 0.5 produces 86.53\%.

We additionally compare ordinary shuffled sampling with strategies that balance the frequency of pedagogical strategies or jointly balance strategies and tutoring turns.

Ordinary shuffled sampling reaches 90.24\% validation accuracy. Strategy-balanced and turn--strategy-balanced sampling achieve 79.80\% and 75.42\%, respectively.

These results favor retaining unique positive winners without an additional margin threshold and training on their observed distribution. This configuration preserves the contextual distribution of successful strategy-conditioned responses during distillation.

\subsection{Interaction-Based Checkpoint Selection}
Finally, we examine checkpoint selection using downstream tutoring performance. We evaluate saved training checkpoints through complete five-turn validation interactions and select the checkpoint using final cumulative accuracy.

On GSM8K, the interaction-selected checkpoint at step 220 achieves 91.92\% validation accuracy, compared with 88.55\% for the minimum-validation-loss checkpoint at step 120. The corresponding mean accuracy across all five tutoring turns increases from 85.72\% to 87.81\%.

The difference highlights the value of selecting the deployed tutor using the downstream objective. We therefore use interaction-based checkpoint selection for the final \method{} tutors.

\section{Response Quality Evaluation}
\label{app:resp-qeval}
We use \texttt{GPT-5.6-Luna} (Luna) as an LLM judge to evaluate a larger sample of student and tutor responses than we can annotate by hand. The GSM8K audit covers 270 initial student replies, 180 student replies during tutoring, and 180 fixed-strategy tutor responses. We obtain human ratings for selected subsets to assess agreement with Luna on the same examples.

\subsection{Student Response Quality}
\label{app:student-resp-qeval}
The student audit evaluates the two-call \texttt{Qwen2.5-32B} simulator. The 270 initial replies cover knowledge gaps, misconceptions, and mixed profiles with two to four problem-relevant difficulties. Another 180 replies are sampled during tutoring with the six fixed strategies. For each affected KC, Luna checks whether the public reply follows the assigned knowledge gap or misconception.

Table~\ref{app:tab:student-quality} reports consistency across all affected KCs and among KCs reached in the response. In the first measure, a KC not reached counts as consistent because its assigned state has not been violated. The entire response requires consistency with every affected KC. Answer agreement checks whether the separately returned answer agrees with the public reply.

\begin{table}[!htb]
\centering
\footnotesize
\setlength{\tabcolsep}{2.5pt}
\renewcommand{\arraystretch}{1.05}
\begin{tabularx}{\columnwidth}{
@{}
>{\raggedright\arraybackslash}X
c
c
c
c
c
@{}}
\toprule
Student reply & $N$ &
\makecell{All\\KCs} &
\makecell{Reached\\KCs} &
\makecell{Full\\response} &
\makecell{Answer\\agree} \\
\midrule
Before tutoring      & 270 & 78.1 & 63.7 & 58.1 & 100.0 \\
After tutor feedback & 180 & 81.8 & 76.1 & 61.1 & 97.8 \\
\bottomrule
\end{tabularx}
\caption{Luna's evaluation of student replies on GSM8K (\%). All KCs and Reached KCs summarize KC-level judgments. Entire response and Answer agreement summarize response-level judgments. $N$ is the number of replies, not the number of KCs.}
\label{app:tab:student-quality}
\end{table}

\paragraph{Knowledge-state changes:} Thirty-one of the 180 tutoring replies follow a recorded state change. Luna judges 29 changes supported by tutor feedback (93.5\%) and 30 restricted to the appropriate KC (96.8\%). These percentages use only turns with an actual change.

\subsection{Tutor Response Quality}
\label{app:tutor-resp-qeval}
Luna evaluates 180 \texttt{Qwen2.5-7B-Instruct} tutor responses, with 30 per strategy and 36 per turn across turns one through five. It checks whether the requested strategy appears, whether it is the main form of assistance, whether the mathematics is correct, and whether the correct final answer is disclosed. Answers to analogous examples do not count as disclosure of the original problem's answer. Table~\ref{app:tab:tutor-quality} reports these judgments separately.

\begin{table}[!htb]
\centering
\small
\setlength{\tabcolsep}{4pt}
\renewcommand{\arraystretch}{1.08}
\begin{tabularx}{\columnwidth}{@{}>{\raggedright\arraybackslash}Xr@{}}
\toprule
Criterion & Luna (\%) \\
\midrule
Requested strategy appears & 96.1 \\
Requested strategy is the main approach & 92.8 \\
Mathematically correct feedback & 95.0 \\
Correct final answer disclosed & 0.0 \\
\bottomrule
\end{tabularx}
\caption{Luna's evaluation of 180 fixed-strategy tutor responses. The strategy-presence and main-approach criteria distinguish including a strategy from using it as the primary assistance.}
\label{app:tab:tutor-quality}
\end{table}

\subsection{Human Evaluation}
\label{app:human-evaluation}

We recruit English-fluent crowd annotators on Prolific for three studies. Each contains 20 selected GSM8K examples from the corresponding Luna audit and three ratings per item. Annotators receive definitions and practice examples. Model identities and Luna's labels are hidden. Two matching votes determine a majority label, and three-way disagreements remain unresolved. Table~\ref{app:tab:human-evaluation} compares Luna and human judgments on the same items. Exact agreement is the fraction assigned the same categorical label.

For student replies, annotators see the problem, assigned knowledge gaps or misconceptions, and public reply. They label each affected KC as consistent, violated, not reached, or unclear. The 20 replies contain 59 KCs. Fifty have a resolved majority, and nine remain unresolved. Of the 31 KCs judged reached with a definite majority, 21 are consistent with the assigned knowledge (67.7\%). The table compares the complete labels on all 50 majority-resolved KCs.

For state changes, annotators review the tutor feedback, student replies, and before-and-after KC states. Twelve selected events contain a change. They assess whether feedback supports the update and addresses the changed KC. The other eight events have no state change, which all human majorities recognize. For tutor responses, annotators see the problem, dialogue, requested strategy definition, and feedback. They separately judge the strategy used, mathematical correctness, and correct final-answer disclosure.

\begin{table*}[!htb]
\centering
\small
\setlength{\tabcolsep}{5pt}
\renewcommand{\arraystretch}{1.10}
\begin{tabularx}{\textwidth}{@{}>{\raggedright\arraybackslash}Xrrr@{}}
\toprule
Judgment & Luna & Human majority & Exact agreement \\
\midrule
\multicolumn{4}{@{}l}{\textit{Student replies: 50 KCs with a resolved human majority}} \\
\quad Consistent with assigned knowledge & 24/50 (48.0\%) & 21/50 (42.0\%) & \multirow{3}{*}{42/50 (84.0\%)} \\
\quad Violates assigned knowledge & 7/50 (14.0\%) & 10/50 (20.0\%) & \\
\quad Relevant step not reached & 19/50 (38.0\%) & 19/50 (38.0\%) & \\
\midrule
\multicolumn{4}{@{}l}{\textit{Knowledge-state changes: 12 changed events}} \\
\quad Feedback supports the change & 11/12 (91.7\%) & 10/12 (83.3\%) & 11/12 (91.7\%) \\
\quad Feedback addresses the changed KC & 11/12 (91.7\%) & 11/12 (91.7\%) & 12/12 (100.0\%) \\
\midrule
\multicolumn{4}{@{}l}{\textit{Tutor feedback: 20 responses}} \\
\quad Requested strategy is the main approach & 19/20 (95.0\%) & 20/20 (100.0\%) & 19/20 (95.0\%) \\
\quad Mathematically correct & 18/20 (90.0\%) & 19/20 (95.0\%) & 19/20 (95.0\%) \\
\quad Correct final answer disclosed & 0/20 (0.0\%) & 0/20 (0.0\%) & 20/20 (100.0\%) \\
\bottomrule
\end{tabularx}
\caption{Luna and human judgments on the same items. Counts give the number assigned each label, with percentages in parentheses. Exact agreement compares the complete categorical judgments. For student replies, it compares all three labels across the 50 KCs with a human majority. The nine unresolved KCs are excluded from this comparison but retained in the rater-agreement analysis.}
\label{app:tab:human-evaluation}
\end{table*}

\paragraph{Interpreting agreement:} Luna and human majorities agree on 84.0\% of resolved student-KC labels, 91.7\% of judgments about support for state changes, and 95.0\% of tutor strategy and correctness judgments. Human pairwise agreement on student-KC labels is 41.8\%. Table~\ref{app:tab:judge-reliability} provides chance-corrected measures for every criterion. All human strategy majorities are positive, and all disclosure labels are negative. The two judges also receive different information. Luna sees the reference solution and additional state information for student judgments. Humans do not. Neither sees private student reasoning. Humans see the requested tutor strategy, while Luna classifies the strategies without that request.

\begin{table*}[!htb]
\centering
\small
\setlength{\tabcolsep}{5pt}
\renewcommand{\arraystretch}{1.08}
\begin{tabularx}{\textwidth}{@{}>{\raggedright\arraybackslash}Xrrr@{}}
\toprule
Criterion & \makecell{Human pairwise\\agreement (\%)} & \makecell{Luna--human\\$\kappa$} \\
\midrule
Student knowledge-state labels & 41.8 & 0.74 \\
Feedback supports the change & 77.8 & 0.62 \\
Feedback addresses the changed KC & 94.4 & 1.0 \\
Requested strategy is the main approach & 93.3 & 0.83 \\
Mathematically correct tutor feedback & 86.7 & 0.64 \\
\bottomrule
\end{tabularx}
\caption{Agreement statistics for the human studies. Unweighted Cohen's $\kappa$ compares Luna with human majorities. For student knowledge-state labels, human agreement uses all 59 KCs and the Luna comparison uses the 50 majority-resolved KCs. A dash indicates an undefined coefficient.}
\label{app:tab:judge-reliability}
\end{table*}

\subsection{Response to Different Feedback}
\label{app:feedback-controls}

We examine whether the student's state changes when feedback addresses its assigned difficulty, rather than merely asking it to reconsider. \citet{dofaithfulness2026} show that simulated students can abandon a misconception after feedback that does not address it. Inspired by their comparison, we evaluate five feedback conditions. Relevant feedback addresses the assigned knowledge gap or misconception. Retry only asks the student to attempt the problem again. Irrelevant feedback supplies information unrelated to the problem. Misaligned feedback addresses a different KC. Answer reveal provides the final answer without explaining the method. In this last condition, we expect the student to report the supplied answer while retaining its KC state, since reporting an answer alone does not establish mastery of the method.

We select 40 GSM8K problem--profile pairs: 20 with a knowledge gap and 20 with a misconception. Each selected problem activates exactly one of the profile's assigned deficits. Both initially correct and initially incorrect responses are included. We apply each feedback condition separately to the same initial response and state, then record the student's answer, public response, and updated KC state. Table~\ref{app:tab:feedback-state} reports changes to the targeted KC.

\begin{table}[!htb]
\centering
\small
\setlength{\tabcolsep}{6pt}
\renewcommand{\arraystretch}{1.08}
\begin{tabular}{@{}lr@{}}
\toprule
Feedback & \makecell{Target KC\\changed} \\
\midrule
Relevant & 40/40 \\
Retry only & 0/40 \\
Irrelevant & 0/40 \\
Misaligned & 0/40 \\
Answer reveal & 0/40 \\
\bottomrule
\end{tabular}
\caption{Recorded KC-state changes under different feedback on GSM8K. Each condition starts from the same initial response and state. Relevant feedback changes only the targeted KC. None of the four control conditions changes the student's state.}
\label{app:tab:feedback-state}
\end{table}

The student's state changes only after relevant feedback. Of the 40 target-state changes, 39 mark the target KC as mastered, while one does not mark the KC as mastered. An LLM judge considers all 40 changes supported by the feedback and local to the targeted KC. Answer revelation raises immediate answer accuracy to 82.5\%, while leaving every KC state unchanged. When the student is subsequently evaluated from that unchanged state without the revealed answer, accuracy is 7.5\%. This comparison distinguishes using the supplied answer from changing the student's represented knowledge.

We additionally assess whether the public responses follow the resulting states. Across the initial response and five feedback branches for each pair, yielding 240 judged responses, consistency is 90.4\% over all affected KCs and 88.9\% over KCs reached during reasoning. The structured answer agrees with the public response in 100.0\% of cases. Thus, the evaluation checks both whether feedback changes the state selectively and whether the student's subsequent reasoning reflects that state.

\subsection{Transfer to New Problems}
\label{app:student-transfer}

A state change after feedback should affect how the student solves subsequent problems. We therefore test whether the updated KC state helps on a new problem after removing the original problem and dialogue.

We use the same 40 GSM8K teaching cases as the feedback comparison, with 20 knowledge gaps and 20 misconceptions. Each case is paired with seven or eight new problems requiring the same target KC and none of the profile's other assigned gaps or misconceptions. This gives 300 distinct test problems covering four KCs. The original teaching problems come from the training split. For each new attempt, the student receives only the new problem and its profile reconstructed from the state before or after feedback. The original problem, dialogue, feedback, and answer are excluded. We use deterministic decoding and evaluate identical states only once.

\begin{table}[!htb]
\centering
\small
\setlength{\tabcolsep}{5pt}
\renewcommand{\arraystretch}{1.08}
\begin{tabularx}{\columnwidth}{@{}Xr@{}}
\toprule
Student state & Accuracy (\%) \\
\midrule
Before feedback & 5.0 \\
After retry only & 5.0 \\
After irrelevant feedback & 5.0 \\
After misaligned feedback & 5.0 \\
After relevant feedback & \textbf{67.7} \\
After answer reveal & 5.0 \\
\bottomrule
\end{tabularx}
\caption{Accuracy on 300 new GSM8K problems using the student's state before or after feedback on a different problem. Control conditions preserve the original state and share the same evaluated answers.}
\label{app:tab:student-transfer}
\end{table}

Relevant feedback raises accuracy on the new problems from 5.0\% to 67.7\% (Table~\ref{app:tab:student-transfer}). The improvement over irrelevant feedback is 62.7 percentage points, with a 95\% confidence interval of $[53.2,71.6]$. The interval resamples the 40 source cases while keeping their target problems together. Accuracy increases from 0.0\% to 70.0\% for knowledge-gap students and from 10.0\% to 65.3\% for misconception students. These results show that the updated state helps the simulated student solve new problems requiring the same KC, even without the earlier dialogue. The test covers four KCs, and the student's model parameters remain fixed.

\newtcblisting{promptbox}[2][]{
  enhanced,
  breakable,
  listing only,
  listing engine=listings,
  colback=black!3,
  colframe=black!22,
  colbacktitle=black!9,
  coltitle=black,
  fonttitle=\bfseries\footnotesize,
  title={#2},
  title after break={#2 (continued)},
  boxrule=0.35pt,
  arc=1pt,
  left=4pt,
  right=4pt,
  top=4pt,
  bottom=4pt,
  before skip=7pt,
  after skip=7pt,
  listing options={
    basicstyle=\ttfamily\footnotesize\color{black},
    breaklines=true,
    breakatwhitespace=false,
    columns=fullflexible,
    keepspaces=true,
    showstringspaces=false,
    tabsize=2
  },
  #1
}

\section{Prompt Templates}
\label{app:prompts}
This appendix provides the prompts used to construct knowledge components (KCs), simulate students, generate tutor feedback, and evaluate responses. Angle brackets and braces mark fields filled in at runtime. Private student information is not shown to the tutor, and the reference solution is not shown to the student.

\subsection{Knowledge Component Construction}
\begin{promptbox}{Reasoning steps from reference solutions}
Extract atomic reasoning observations from a complete worked solution.
For every supplied authored line, return exactly one line record with status
``represented``, ``non_reasoning``, or ``failed``. Split compound lines and
preserve all reasoning opportunities, including prose-only transformations.
Auxiliary context may clarify terse notation, but evidence_text must always be
an exact substring of the corresponding authored line.

Do not create an observation merely for reporting an answer letter, naming an
answer option, or matching an already computed value to a multiple-choice option.
For example, ignore ``answer: c`` and ``therefore option d``. When a line contains
mathematical reasoning followed by an answer label, represent the mathematical
reasoning and omit only the answer-label fragment. Candidate substitution or
option elimination remains reasoning when it performs a mathematical test.

Split compound lines. Each observation must contain exactly one transformation and:
evidence_text (an exact substring of authored_line), evidence_occurrence (zero-based),
operation (open string), optional expression and result, inputs [{role,value}],
optional output {role,value}, applicability_conditions, goal, transformation, and
one short topic-free description. Roles must be snake_case variable names. Do not
use story objects, names, numerical values,
or particular units in conditions, roles, goals, transformations, or descriptions.
Preserve direction and the unknown role. Operation is metadata, not a KC.

Use each local line_index exactly once and preserve reasoning order. Return only JSON:
{"lines":[{"line_index":0,"status":"represented|non_reasoning|failed",
"reason":"","observations":[{"evidence_text":"...","evidence_occurrence":0,
"operation":"...","expression":"","result":"",
"inputs":[{"role":"...","value":"..."}],
"output":{"role":"...","value":"..."},
"applicability_conditions":["..."],"goal":"...",
"transformation":"natural-language explanation",
"description":"...","issue":"..."}]}]}
\end{promptbox}

\begin{promptbox}{Identifying knowledge components}
Partition ONE coarse cluster into KC families and atomic leaf KCs.
The cluster is only a context and may contain several families. A family is a broad
reusable structure such as equal_groups. A leaf KC is one parameterized production
rule described by natural-language semantics and generalized input/output roles.
Do not propose an operation or symbolic equation at this stage. Contextual objects
and units are instances, not separate KCs.
For example, total cost and total objects can both instantiate equal_groups.find_total.
Different unknown roles or inverse directions may remain different leaves.
List input_roles in the order corresponding to each member observation's inputs.

Each observation has a local integer index. Assign every index exactly once to
one leaf KC or unmatched. Return JSON:
{"families":[{"name":"snake_case","definition":"...","kcs":[{
"name":"family.directional_capability","definition":"...",
"applicability_conditions":["..."],"input_roles":["..."],
"output_role":"...",
"observation_indices":[0,1],"contrast":"..."}]}],
"unmatched":[{"observation_index":2,"reason":"..."}]}
\end{promptbox}

\begin{promptbox}{Combining related knowledge components}
Propose high-confidence synonym merges among preliminary KC families.
Families with the same normalized name are already grouped. Mention only groups
whose broad reusable capabilities are equivalent despite different names. Every
family index may appear in at most one merge; unmentioned families remain separate.
Never merge merely because both use the same arithmetic operation. Return JSON:
{"family_merges":[{"family_indices":[0,1],
"canonical_name":"snake_case","definition":"..."}]}
\end{promptbox}

\begin{promptbox}{Removing overlapping knowledge components}
Finalize the leaf KCs for ONE already-consolidated family.
You are shown every local leaf candidate in this family. Partition every index
exactly once into final leaf KCs. Merge semantically equivalent directional
capabilities even when local names or role names differ. Keep different goals,
unknown roles, or inverse directions separate.

The user message includes required_leaf_indices. Before returning, verify that
the union of leaf_indices equals this list exactly, with no omission or repetition.

Return one object for every final leaf KC, including singleton leaves. Every
canonical_name must be unique within this family. Make this decision using only
semantic capability. Do not return operations, input roles, output roles, or
equations.
Return JSON:
{"leaf_kcs":[{"leaf_indices":[0,1],
"canonical_name":"family.capability","definition":"...",
"applicability_conditions":["..."],"contrast":"..."}]}
\end{promptbox}

\subsection{Problem Annotation}
\begin{promptbox}{Assigning KCs to solution steps}
You map authored solution lines to a closed knowledge-component library.

For every source line:
1. identify zero, one, or multiple atomic reasoning steps;
2. write a short description without copying story-specific numbers;
3. assign exactly one supplied KC name to each reasoning step, or "unmatched";
4. if the line contains no reasoning, mark it non_reasoning;
5. if it cannot be analyzed, mark it failed.

Do not invent KC names. Do not add dependencies, equations, numerical results,
or solution steps absent from the supplied lines.

Return one JSON object:
{
  "steps": [
    {
      "source_order": 1,
      "atomic_order": 1,
      "description": "short reasoning capability application",
      "kc_name": "supplied KC name or unmatched"
    }
  ],
  "dispositions": [
    {
      "source_order": 1,
      "status": "represented|non_reasoning|failed",
      "reason": "required when no step is represented"
    }
  ]
}
\end{promptbox}

\begin{promptbox}{Input for KC annotation}
{
  "problem": "<problem>",
  "gold_solution": "<reference solution>",
  "source_lines": [
    {"source_order": 1, "text": "<authored solution line>"}
  ],
  "frozen_kcs": [
    {
      "name": "<KC name>",
      "definition": "<KC definition>",
      "positive_examples": ["<positive example>"],
      "near_miss_examples": ["<near-miss example>"]
    }
  ]
}
\end{promptbox}

\subsection{Misconception Construction}
\begin{promptbox}{Incorrect beliefs for mathematics}
You design controlled synthetic learner beliefs
for mathematics-tutoring research. Given one correct knowledge component,
propose exactly {candidate_count} distinct and systematic novice beliefs.

Each belief must:
- be one atomic, reusable reasoning error rather than an arithmetic slip;
- plausibly guide work across different problems using the KC;
- state when the learner applies it;
- give a worked numerical example that follows the believed procedure;
- use topic-free role meanings instead of story objects;
- produce public reasoning distinguishable from the other candidates;
- avoid vague claims such as "the learner does not understand";
- avoid arbitrary operator substitution unless a coherent belief explains it.

Write belief_statement from the learner's perspective without calling it
wrong, buggy, or a misconception. wrong_rule should be a compact symbolic
relation using the supplied role names when possible. contrast is private
audit metadata explaining exactly how the belief differs from the correct KC.
Do not create identifiers; code assigns them after validation.
\end{promptbox}

\begin{promptbox}{Incorrect beliefs for other subjects}
You design controlled synthetic learner
beliefs for tutoring research. Given one correct knowledge or reasoning skill,
propose exactly {candidate_count} distinct and systematic novice beliefs.

Each belief must:
- be one coherent, reusable false belief rather than a random mistake;
- plausibly guide answers to different questions that use the skill;
- cover the broad skill rather than a rare wording or narrow subtype;
- state precisely when the learner applies it;
- include a short worked example of applying the false belief;
- lead the learner to a concrete, predictably wrong answer on typical
  multiple-choice questions for the skill;
- produce public reasoning distinguishable from the other candidates;
- avoid vague claims such as "the learner does not understand";
- remain within the supplied skill instead of inventing another topic.

Write belief_statement from the learner's perspective without calling it
wrong, buggy, or a misconception. wrong_rule should concisely state the false
general rule in words; use a symbolic relation only when the skill itself is
symbolic. contrast is private audit metadata explaining exactly how the belief
differs from the correct skill. Do not create identifiers; code assigns them
after validation.
\end{promptbox}

\subsection{Stateful Student Simulator}
For the initial attempt, one call generates private reasoning and another produces the public student reply. After tutor feedback, the first call updates the KC state and the second generates the student reply using that state.

\begin{promptbox}{Initial reasoning with a knowledge gap}
Privately attempt the original
problem using only the student's current understanding supplied below. Treat
that understanding as authoritative rather than using stronger mathematics.

Before calculating, write a capability_plan that places the problem's required
calculations in their actual execution order and identifies every supplied
method governing them. Check every supplied method; do not plan one and forget
another.

Find the earliest calculation in that global order governed by ANY unavailable
method. That is the single stopping boundary. Stop BEFORE writing, simplifying,
or evaluating it, and end the attempt there. Do not continue with another
branch, a hypothetical example, or any later calculation. Even obvious
arithmetic is forbidden when it instantiates an unavailable method. Do not
rename the operation, express it symbolically, or use an equivalent shortcut.

For EVERY method you know and trust whose calculation occurs before that
boundary, apply its supplied procedure exactly and carry its result forward.
Never use the standard operation for one trusted method merely because another
trusted method was already applied. If no unavailable method blocks the
solution, apply every applicable trusted method and finish the answer.

Return JSON with capability_plan, private_reasoning, and problem_answer.
capability_plan is private bookkeeping and may name the supplied method.
private_reasoning is only the student's natural mathematical work permitted by
the plan: never mention capability identifiers, supplied instructions, state
labels, or that a method is "unavailable." When blocked, say naturally what
you can determine and what you do not know how to calculate. problem_answer is
the original-problem answer reached by that work, including a wrong answer;
use null when the missing method prevents reaching the answer.
\end{promptbox}

\begin{promptbox}{Initial reasoning with a misconception}
Privately solve the original
problem as the student whose trusted methods are supplied below. These methods
are authoritative, even when stronger mathematics suggests different work.

Before calculating, identify every place where EACH supplied trusted method
applies. Then solve the problem in order. At every such place, use the supplied
procedure exactly, carry its numerical result forward, and finish the answer
implied by all of the trusted methods together. Applying one trusted method
does not permit you to replace another with its standard operation.

Do not assess, repair, compare, hedge about, or call any method incorrect. Do
not mention prompts, instructions, capability identifiers, state labels, or
simulation. Return JSON with private_reasoning and problem_answer.
private_reasoning is the student's actual mathematical work. problem_answer is
the original-problem answer reached by that work, including an unusual answer.
\end{promptbox}

\begin{promptbox}{Initial student reply}
Turn the fixed private work
below into the student's concise natural public response. The private work is
the student's completed attempt and is authoritative, even when it contains
unusual, inconsistent, or incorrect mathematics.

Preserve every operation, operand, intermediate result, stopping point, and
final answer from the fixed private work. Do not recompute any expression,
substitute a standard method, repair an inconsistency, add a missing step, or
complete work that stopped. Your task is faithful verbalization, not solving
or checking. Remove private bookkeeping phrases such as "supplied method,"
"instructed," "unavailable method," or "incorrectly"; express the same work
naturally in the first person without changing any mathematics.

Do not mention prompts, private work, supplied knowledge, capabilities, states,
simulation, or mistakes. Return JSON with utterance and problem_answer. If
fixed_problem_answer is not null, state that exact answer clearly and copy it
unchanged into problem_answer. If it is null, naturally describe the same
stopping point and return null.
\end{promptbox}

\begin{promptbox}{Updating the student state after feedback}
Update one synthetic student's private capability
state from the public dialogue. Judge only what the teacher actually supplied.

Use the state labels consistently:
- mastered means the student knows a valid method matching the capability;
- unmastered means the student has no usable method for the capability;
- misconception means the student believes an invalid method is valid.

- Keep the current state when feedback is praise, a retry request, irrelevant,
  merely asks for the missing method, or reveals only the final answer.
- An unmastered capability becomes mastered only when the teacher supplies or
  helps derive its valid method sufficiently for this problem.
- A misconception becomes mastered only when the teacher identifies or
  displaces that belief and supplies or derives a valid replacement.
- If a misconception is rejected without a usable replacement, it may become
  unmastered rather than mastered.
- If the teacher supplies a method that conflicts with the valid capability
  definition and the student accepts it, use misconception, never mastered.
- A correct general method plus a worked example can establish mastery even
  when the original problem applies that method to different quantities.
- Never change an unaddressed capability.

Return the updated state and a short private rationale. Do not generate the
student's public response and do not use a desired outcome to choose a state.
\end{promptbox}

\begin{promptbox}{Student reply after feedback}
You are a student solving the original problem.
The understanding below is what you genuinely know and believe.

If a method is unavailable, do the parts you understand and stop before that
method. Do not reconstruct it or use an equivalent shortcut.

If a method is supplied as your belief, you believe it is correct and always
apply it when its stated situation occurs. This is simply how you do math.
Use it naturally in your working, carry its result through the problem, and
give that resulting answer. Never mention the rule, describe it as an error,
silently replace it, or compute an alternative expert answer.

Return JSON with utterance and problem_answer. utterance is your natural public
work on the original problem. problem_answer is exactly the original-problem
answer stated in the utterance, even when unusual or wrong; use null only if
the original problem remains unresolved.
\end{promptbox}

\begin{promptbox}{Input for initial student reasoning}
ORIGINAL PROBLEM:
<problem>

CURRENT UNDERSTANDING:
<problem-relevant KC definitions and the student's current gap or misconception descriptions>

Now privately carry out the attempt from this understanding.
\end{promptbox}

\begin{promptbox}{Input for the initial student reply}
{
  "original_problem": "<problem>",
  "fixed_private_reasoning": "<private reasoning from the first call>",
  "fixed_problem_answer": "<answer from the first call, or null>"
}
\end{promptbox}

\begin{promptbox}{Input for updating the student state}
{
  "public_problem": "<problem>",
  "capabilities": [
    {
      "kc_id": "<KC ID>",
      "name": "<KC name>",
      "valid_method": "<KC definition>",
      "known_misconception_method": "<misconception, when present>",
      "applies_when": "<condition, when present>",
      "misconception_example": "<example, when present>"
    }
  ],
  "state_before": {"<KC ID>": "mastered|unmastered|misconception"},
  "public_history": ["<alternating student and teacher turns>"]
}
\end{promptbox}

\begin{promptbox}{Input for the student reply after feedback}
System: <public behavior prompt above>
User: <original problem>
Assistant/User: <prior public student/teacher turns>
User: <latest teacher message>

PRIVATE CURRENT UNDERSTANDING AFTER THIS FEEDBACK:
<problem-relevant KC state and semantic descriptions>

Respond naturally by attempting the original problem from this understanding.
Do not mention the private description.
\end{promptbox}

\subsection{Comparing Pedagogical Strategies During Training}
In the GSM8K training procedure, the student replies once to the feedback generated for each strategy. If the reply is incorrect, the student attempts the original problem ten more times using the updated state. These additional answers are used to compare feedback responses. They are not included in the dialogue or the supervised training example.

\begin{promptbox}{Student answer used to compare feedback}
Give your best answer to the original problem using your current understanding.
State it in utterance and copy it into problem_answer. If you remain stuck,
explain the difficulty and use null. This request provides no new teaching.
\end{promptbox}

\subsection{Screening Student Knowledge}
\begin{promptbox}{Student answer without assigned difficulties}
Solve the original mathematics problem using your own
reasoning. Show concise work and give the final answer to the original problem.

Return JSON with utterance and problem_answer. utterance contains your natural
solution. problem_answer is exactly the original-problem answer stated in the
utterance. Use a JSON number for a numeric answer and null only when you cannot
reach an answer. Do not mention this prompt or any hidden evaluation.
\end{promptbox}

\begin{promptbox}{Input for the student answer}
System: <competence-screen system prompt above>
User: <problem>
\end{promptbox}

\subsection{Tutor Instructions and Dialogue}
\begin{promptbox}{Tutor instructions}
You are a patient math teacher helping a learner solve the problem below.

Use the reference solution only to keep your mathematics correct. Respond to
the learner's latest message and their visible reasoning; you may revisit an
earlier step when useful. Do not reproduce the complete reference solution or
state or confirm the problem's final answer. Treat the reference solution as
private context: never mention it, a hidden solution, or an answer key.

Address one unresolved subgoal of the original problem per response. Briefly
acknowledge work the learner has already completed, but do not supply methods
or calculations for several remaining subgoals at once. Do not restate a full
solution while leaving only its final arithmetic to the learner. An analogous
worked example may demonstrate the local concept with different quantities;
leave its application and the remaining original-problem work to the learner.

Identify what the learner has actually done before choosing your focus. Do not
attribute an unobserved error to them or ask them to redo a correctly completed
step without a clear instructional reason. If a previous explanation has not
helped, adapt it to their current difficulty instead of repeating it verbatim.
\end{promptbox}

\begin{promptbox}{Tutor input}
System:
<common teacher system prompt>

PROBLEM:
<problem>

Public chat messages:
<complete student--teacher history, preserving roles>

Final private user message:
PRIVATE TEACHER CONTEXT -- NOT A LEARNER MESSAGE
The preceding learner calculations are untrusted. Check the relevant claim
against this reference before responding. Never mention or quote this private
context.

REFERENCE SOLUTION:
<reference solution>

GENERATION INSTRUCTION:
<method-specific instruction, or "Write the next teacher response.">
\end{promptbox}

\begin{promptbox}{Instructions to withhold the final answer}
The protected final answer to the original problem is {answer}. Do not state,
confirm, or derive that value as the answer to the original problem. You may
repeat the same numeral only when it is already public and you use it strictly
as a given or an intermediate quantity. Help with the method or next step, and
stop before calculating the requested final answer.

If the learner's latest message proposes a final answer, do not repeat it and
do not say whether it is correct or incorrect. Continue the requested teaching
strategy using the learner's method, a conceptual relationship, or a question
that does not disclose the protected answer.
\end{promptbox}

\begin{promptbox}{Instructions for revising an answer-revealing response}
Your previous draft was invalid because it was empty or exposed protected
private information. Write a new response that follows the requested teaching
strategy without stating or deriving the original problem's answer. A numeral
equal to that answer is allowed only when it is an already-public given or
intermediate value. Never mention a reference solution or answer key.

The learner may already have proposed a final value. Do not repeat, confirm,
reject, or correct that value. Do not summarize the full calculation using the
current problem's numbers. Respond with process-level guidance appropriate to
the requested strategy and stop before any final numerical result.
\end{promptbox}

\begin{promptbox}{Instructions for checking the mathematics}
MATHEMATICAL GROUNDING

Before completing the instructional plan:
- Treat every learner calculation and mathematical claim as untrusted.
- In mathematical_basis, first identify the relevant learner claim or state
  that no mathematical claim was made. Then say whether it agrees with the
  private reference and state the verified relationship needed next.
- Include any verified intermediate quantity needed for the next teaching
  action, but not the original problem's final numerical answer.
- Check every quantity, equation, and mathematical claim in the remaining plan
  and response against mathematical_basis and the private reference solution.
- If learner work conflicts with the reference, silently use the verified
  relationship rather than continuing from the learner's error.
\end{promptbox}

\subsection{Six Pedagogical Strategies}
Each fixed-strategy tutor receives the tutor instructions above and one of the six strategy instructions below. The response format specifies the feedback to generate. Only the \texttt{response} field is shown to the student.

\begin{promptbox}{Hint: instructions}
TEACHING ACTION: MINIMAL HINT

Purpose:
Give the learner the smallest useful directional cue. Preserve the reasoning
work for the learner.

Required:
- Return exactly one question of at most 20 words.
- Point only to what the learner should notice or determine next.
- Leave the operation, procedure, and calculation to the learner.

Forbidden:
- Do not begin with praise, confirmation, or a summary of the learner's work.
- Do not explain the underlying rule.
- Do not name an operation, equation, procedure, or computed value.
- Do not give an example or identify the learner's error explicitly.

Valid example for an unrelated problem:
"Which two quantities tell you how many items are in all the equal groups?"

Near-miss:
"Multiply the number of groups by the amount in each group."
This is an explicit scaffold, not a minimal hint.

Complete the private instructional plan described after this action.
\end{promptbox}

\begin{promptbox}{Hint: response format}
PRIVATE INSTRUCTIONAL PLAN

Return exactly one JSON object with keys in this order:
{
  "mathematical_basis": "the verified next relationship and any needed verified intermediate quantity, without the final numerical answer",
  "conceptual_focus": "the role or relationship the learner should notice",
  "question": "one question of at most 20 words",
  "response": "exactly the same text as question"
}

Derive mathematical_basis from the private reference solution rather than
assuming the learner's intermediate values are correct. Then fill the strategy
fields and response last. Only response will be shown to the student. It must
naturally direct attention to conceptual_focus while leaving the operation and
execution to the learner.

Concrete example for an unrelated problem:
{
  "mathematical_basis": "An equal-groups total relates the group count to the equal amount in each group.",
  "conceptual_focus": "the two roles that define the equal groups",
  "question": "Which quantities tell you the number of groups and the amount in each?",
  "response": "Which quantities tell you the number of groups and the amount in each?"
}
\end{promptbox}

\begin{promptbox}{Step scaffold: instructions}
TEACHING ACTION: STEP SCAFFOLD

Purpose:
Provide the structure of the learner's next step while leaving its execution
and the final answer to the learner.

Required:
- Return one or two sentences.
- Supply one explicit operation, unevaluated equation, or ordered procedure
  for exactly one next subgoal in the current problem.
- End by asking the learner to carry out that single step.

Forbidden:
- Do not begin with generic praise or summarize all prior work.
- Do not calculate the requested value or state the final answer.
- Do not switch to a different example.
- Do not respond with only a vague question.

Valid example for an unrelated problem:
"Write total = number of bags times marbles per bag, substitute the two given
quantities, and then evaluate it."

Near-miss:
"What quantity should you find next?"
This is a minimal hint because it identifies a subgoal but supplies no method.

Complete the private instructional plan described after this action.
\end{promptbox}

\begin{promptbox}{Step scaffold: response format}
PRIVATE INSTRUCTIONAL PLAN

Return exactly one JSON object with keys in this order:
{
  "mathematical_basis": "the verified next relationship and any needed verified intermediate quantity, without the final numerical answer",
  "subgoal": "the single next subgoal",
  "step_structure": "the explicit operation, unevaluated equation, or ordered procedure",
  "learner_request": "what the learner should carry out",
  "response": "a natural response expressing the preceding fields"
}

Derive mathematical_basis from the private reference solution rather than
assuming the learner's intermediate values are correct. Then fill the strategy
fields and response last. Only response will be shown to the student. It must
naturally express the subgoal and step_structure and add no other teaching
action.

Concrete example for an unrelated problem:
{
  "mathematical_basis": "Equal groups relate a group count to one repeated amount.",
  "subgoal": "Represent the total across the equal groups.",
  "step_structure": "total = number of groups * amount per group",
  "learner_request": "Substitute the two known quantities and evaluate it.",
  "response": "Represent the total as number of groups times amount per group. Substitute the two known quantities and evaluate it."
}
\end{promptbox}

\begin{promptbox}{Explicit explanation: instructions}
TEACHING ACTION: EXPLICIT EXPLANATION

Purpose:
Teach the single conceptual or mathematical relationship needed at the
learner's current impasse.

Required:
- Return two or three declarative sentences and use no question marks.
- State what one relevant relationship means, when it applies, and why it
  fits the learner's situation.

Forbidden:
- Do not tell the learner what procedural step to perform next.
- Do not begin with generic praise or summarize every step completed so far.
- Do not solve the full current problem or state its final answer.
- Do not introduce and solve an analogous problem.

Valid example for an unrelated problem:
"Equal-group situations contain the same amount repeatedly. Multiplication
represents that repetition because it combines one equal amount per group."

Near-miss:
"What operation represents equal groups?"
This asks for recall instead of explicitly explaining the relationship.

Complete the private instructional plan described after this action.
\end{promptbox}

\begin{promptbox}{Explicit explanation: response format}
PRIVATE INSTRUCTIONAL PLAN

Return exactly one JSON object with keys in this order:
{
  "mathematical_basis": "the verified next relationship and any needed verified intermediate quantity, without the final numerical answer",
  "relationship": "the single relationship to teach",
  "when_it_applies": "the condition under which it applies",
  "why_it_fits": "why it fits the learner's current situation",
  "response": "a natural declarative explanation of those three fields"
}

Derive mathematical_basis from the private reference solution rather than
assuming the learner's intermediate values are correct. Then fill the strategy
fields and response last. Only response will be shown to the student. It must
naturally explain the planned relationship, when it applies, and why it fits.

Concrete example for an unrelated problem:
{
  "mathematical_basis": "An equal-groups total repeats one equal amount for every group.",
  "relationship": "Multiplication represents repeated equal amounts.",
  "when_it_applies": "It applies when every group contains the same amount.",
  "why_it_fits": "The shelves each hold an equal number of books.",
  "response": "Multiplication represents repeated equal amounts, so it applies whenever every group contains the same amount. Here, the shelves each hold an equal number of books."
}
\end{promptbox}

\begin{promptbox}{Worked example: instructions}
TEACHING ACTION: WORKED EXAMPLE

Purpose:
Demonstrate the relevant reasoning on a short analogous problem and connect
that reasoning to the learner's problem.

Required:
- Return exactly three parts: an analogous mini-problem, its completed
  calculation and result, and one sentence connecting its relationship to the
  learner's problem.
- Use different objects and different values from the current problem.

Forbidden:
- Do not begin with praise or continue solving the current problem.
- Do not calculate any step of the learner's current problem.
- Do not state the learner's final answer.
- Do not merely suggest that the learner imagine an example.
- Do not end by asking the learner to solve the analogous example.

Valid example for an unrelated problem:
"For example, suppose 3 shelves hold 4 books each. The calculation is
3 times 4 = 12 books. This uses the same equal-groups relationship as your
problem."

Near-miss:
"Imagine some shelves with books and use the same idea."
This mentions an example but does not work one through.

Complete the private instructional plan described after this action.
\end{promptbox}

\begin{promptbox}{Worked example: response format}
PRIVATE INSTRUCTIONAL PLAN

Return exactly one JSON object with keys in this order:
{
  "mathematical_basis": "the verified next relationship and any needed verified intermediate quantity, without the final numerical answer",
  "analogous_problem": "a different short problem with different values",
  "worked_calculation": "its completed calculation and result",
  "connection": "how its relationship transfers to the learner's problem",
  "response": "a natural worked example expressing those three fields"
}

Derive mathematical_basis from the private reference solution rather than
assuming the learner's intermediate values are correct. Verify the analogous
calculation arithmetically. Then fill response last. Only response will be
shown to the student. It must naturally include the different problem,
completed calculation, and connection without calculating the learner's
problem.

Concrete example for an unrelated problem:
{
  "mathematical_basis": "An equal-groups total repeats one equal amount for every group.",
  "analogous_problem": "Suppose three shelves hold four books each.",
  "worked_calculation": "The total is 3 * 4 = 12 books.",
  "connection": "This uses the same equal-groups relationship as the learner's problem.",
  "response": "Suppose three shelves hold four books each. The total is 3 * 4 = 12 books. This uses the same equal-groups relationship as your problem."
}
\end{promptbox}

\begin{promptbox}{Self-explanation: instructions}
TEACHING ACTION: SELF-EXPLANATION PROMPT

Purpose:
Make the learner explain the meaning or justification of their own visible
reasoning rather than receive new instruction.

Required:
- Return exactly one concise question.
- Refer to one step, relation, assumption, or proposed approach visible in the
  learner's messages and ask for its meaning or justification.

Forbidden:
- Do not begin with praise or summarize the learner's work.
- Do not supply the relationship the learner is meant to explain.
- Do not ask only what operation or value comes next.
- Do not calculate or correct the step for the learner.

Valid example for an unrelated problem:
"How does the equation you wrote represent the equal groups you described?"

Near-miss:
"Which operation should you use?"
This is a hint or recall question, not a request to explain reasoning.

If the learner has not shown a justification, ask them to explain the meaning
of their current or intended approach. Complete the private instructional plan
for every history.
\end{promptbox}

\begin{promptbox}{Self-explanation: response format}
PRIVATE INSTRUCTIONAL PLAN

Return exactly one JSON object with keys in this order:
{
  "mathematical_basis": "the verified next relationship and any needed verified intermediate quantity, without the final numerical answer",
  "claim_to_explain": "the learner's visible or intended approach, relationship, or assumption",
  "explanation_goal": "the meaning or justification the learner should articulate",
  "question": "one question eliciting that meaning or justification",
  "response": "exactly the same text as question"
}

Derive mathematical_basis from the private reference solution rather than
assuming the learner's intermediate values are correct. Identify what the
learner should explain about their visible or intended approach. Then write
response last. Only response will be shown to the student. It must elicit the
planned meaning or justification rather than merely request the next result.

Concrete example for a history containing "I multiplied by the bags":
{
  "mathematical_basis": "An equal-groups total repeats the amount in one bag for every bag.",
  "claim_to_explain": "using multiplication for the bag relationship",
  "explanation_goal": "why repetition across equal bags is represented by multiplication",
  "question": "Why does multiplying by the number of bags represent this situation?",
  "response": "Why does multiplying by the number of bags represent this situation?"
}
\end{promptbox}

\begin{promptbox}{Contrastive feedback: instructions}
TEACHING ACTION: CONTRASTIVE ERROR FEEDBACK

Purpose:
Contrast an incorrect or incomplete relationship with the relationship the
problem requires.

Required:
- Return two or three sentences.
- State a concrete incorrect relationship or procedure. It must describe what
  someone would actually do, rather than merely omit the correct step.
- If that error is visible, attribute it to the learner. Otherwise, introduce
  it honestly as a possible error without claiming the learner made it.
- Explicitly contrast what the incorrect relationship represents with what
  the problem requires, then ask the learner to revise the relationship.
- In the public response, state the incorrect relationship before giving the
  wrong-versus-right contrast. Do not replace either part with praise.

Forbidden:
- Do not praise or confirm correct work instead of identifying an error.
- Do not attribute an error to the learner unless they expressed it.
- Do not use "not doing the correct step" as the incorrect alternative.
- Do not merely say the answer is wrong.
- Do not state or calculate the final answer.

Valid example for an unrelated problem:
"You added the group count to the group size. Addition uses each quantity once,
but this problem repeats the group size for every group; revise your setup to
represent that repetition."

Near-miss:
"Multiply by the number of days and try again."
This supplies a scaffold but gives no incorrect relationship to contrast.

Complete the private instructional plan for every history.
\end{promptbox}

\begin{promptbox}{Contrastive feedback: response format}
PRIVATE INSTRUCTIONAL PLAN

Return exactly one JSON object with keys in this order:
{
  "mathematical_basis": "the verified next relationship and any needed verified intermediate quantity, without the final numerical answer",
  "error_source": "observed or hypothetical",
  "incorrect_relationship": "a concrete affirmative wrong relationship or procedure",
  "required_relationship": "what the problem instead requires",
  "contrast": "what the wrong relationship represents versus what the required relationship represents",
  "revision_request": "what the learner should reconsider",
  "response": "a natural contrastive response expressing those fields"
}

Derive mathematical_basis from the private reference solution rather than
assuming the learner's intermediate values are correct. Use observed only when
the public dialogue contains an actual mathematical error. Otherwise use a
genuinely incorrect hypothetical relationship and do not attribute it to the
learner. The incorrect relationship must be a positive alternative such as an
incorrect operation or equation; "not using the correct relationship" is not
an alternative. Then write response last. Only response will be shown to the
student. It must naturally state the wrong relationship, contrast its meaning
with the required relationship, and request revision.

Concrete example with an observed error in "I added 3 and 4":
{
  "mathematical_basis": "An equal-groups total repeats the amount in one group for every group.",
  "error_source": "observed",
  "incorrect_relationship": "Add the group count to the group size.",
  "required_relationship": "Equal groups repeat the group size once per group.",
  "contrast": "Addition uses each quantity once, whereas multiplication repeats the group size for every group.",
  "revision_request": "Revise the setup to represent that repetition.",
  "response": "You added the group count and group size, which uses each quantity once. That does not repeat the group size for every group. Equal groups require one copy per group, so revise your setup to represent that repetition."
}

Concrete example with a hypothetical error after "I am stuck":
{
  "mathematical_basis": "An equal-groups total repeats the amount in one group for every group.",
  "error_source": "hypothetical",
  "incorrect_relationship": "One possible mistake would be to add the group count to the group size.",
  "required_relationship": "The group size must occur once for every group.",
  "contrast": "Addition uses each quantity once, whereas multiplication repeats the group size for every group.",
  "revision_request": "Revise the setup so it represents repeated equal groups.",
  "response": "One possible mistake is adding the group count to the group size. That uses each quantity once, whereas this situation repeats the group size once for every group. Revise the setup so it represents that repetition."
}
\end{promptbox}

\subsection{Prompt Selector and AdaTutor}
\begin{promptbox}{Prompt Selector instructions}
Choose the single teaching strategy most useful for the learner's current
public dialogue, then write the teacher response using that strategy.

Allowed strategies:
- MINIMAL_HINT: cue what to notice or determine without supplying the method;
- STEP_SCAFFOLD: supply an operation, equation, or procedure, not its result;
- EXPLICIT_EXPLANATION: declaratively teach one needed relationship;
- WORKED_EXAMPLE: solve an analogous example with different objects and values;
- SELF_EXPLANATION_PROMPT: ask why or how a visible learner step works;
- CONTRASTIVE_ERROR_FEEDBACK: contrast a visible error with the correct relation.

Do not collapse every action into a question. Use the defining form of the
selected strategy. Select SELF_EXPLANATION_PROMPT only when the learner has
shown reasoning to explain. Select CONTRASTIVE_ERROR_FEEDBACK only when the
learner has shown a specific erroneous method or relation to contrast.

Return exactly one JSON object:
{
  "selected_strategy": "one allowed strategy",
  "response": "clean public teacher response with no strategy label"
}
\end{promptbox}

The \textsc{Prompt Selector} receives the tutor instructions and chooses a strategy before generating feedback. AdaTutor receives the same tutoring context and generates the strategy tag and feedback together. The tag is kept for analysis and removed before the feedback is shown to the student.

\begin{promptbox}{AdaTutor output format}
[<STRATEGY>] <teacher response>
\end{promptbox}

\subsection{TutorRL and SocraticLM}
\begin{promptbox}{TutorRL}
You are tasked with being a teacher and helping a student with a math problem.

You must not reveal the answer to the problem to the student at any point in time.
Your task is to guide the student to have a complete understanding of the problem.
Even if the student is already able to solve the problem, you should help them understand and improve the solution so that they get as high of a grade as possible.

If possible, do not respond with overly long responses to the student.

You can end a conversation by writing <end_of_conversation>, please try to end conversations as soon as they are finished instead of prolonging them if not needed. But do not end them prematurely either.

Here is the math problem:
{problem}
\end{promptbox}

\begin{promptbox}{SocraticLM}
You are a Socratic teacher, please guide me to solve the [Problem] with heuristic questions based on the following information.
[Problem]{problem} [Answer] {answer} [Analysis] {analysis}
\end{promptbox}

\subsection{LLM Evaluation of Student and Tutor Responses}
\begin{promptbox}{Student response quality instructions}
You are auditing a prompt-controlled synthetic math
student. Judge the supplied response against its declared capabilities, not
against mathematical correctness or a model of human learning. Read the ENTIRE
public_history_before and latest teacher_message before judging support.

State meanings:
- mastered: the student may use the capability's correct method;
- unmastered: the student cannot independently use or reconstruct that method;
- misconception: the student believes the supplied WRONG belief is correct.
  Before relevant teacher correction, using that wrong belief is FAITHFUL.
  Using the corresponding correct method instead is UNFAITHFUL when the wrong
  belief's stated condition applies and the capability is actually reached.

First judge EACH capability whose state_before is unmastered or misconception:
- reached is true when the response arrives at a calculation governed by that
  capability. It is false when the response stops earlier or the condition
  genuinely does not arise.
- faithful is true for an unreached capability. For a reached unmastered
  capability, it is true only when the response stops before independently
  using it. For a reached misconception, it is true only when the assigned
  wrong belief is used, unless relevant teacher feedback supplied a valid
  replacement.
- Return exactly one capability_judgments row for every active KC ID, even when
  it is not reached. Never return rows for mastered capabilities.

Then judge the whole response:
1. response_consistent_with_state is true only when every active capability is
   faithful. This is the strict joint metric.
2. Relevant mathematical information in the preceding public teacher history
   may support current work without changing stored state.
3. A declared state change requires relevant guidance and must be local.
   Praise, retry requests, unrelated content, and a bare final answer do not
   justify acquisition. Initial responses have no feedback and retain state.
4. problem_answer must equal the original-problem answer expressed in the
   utterance. Null is appropriate when no original answer is stated.
5. Hidden simulator language includes persona, cognitive profile, KC IDs,
   hidden state, injected misconception, or claims of role-playing.

Return exactly one JSON object with one judgment per supplied case:
{
  "judgments": [
    {
      "case_id": "copied case ID",
      "state_transition_supported": true,
      "state_transition_local": true,
      "response_consistent_with_state": true,
      "answer_matches_utterance": true,
      "hidden_simulator_language": false,
      "capability_judgments": [
        {
          "kc_id": "KC copied from state_before",
          "reached": true,
          "faithful": true,
          "rationale": "brief concrete evidence"
        }
      ],
      "rationale": "brief whole-response evidence"
    }
  ]
}
\end{promptbox}

\begin{promptbox}{Tutor response quality instructions}
You are auditing which one-turn instructional actions a math-teacher response
realizes. Classify the response from its content without guessing which action
its generator was asked to use. A natural response may realize more than one
action; record every action substantially present.

Actions:
- MINIMAL_HINT: a concise question pointing to what to notice or determine next
  while leaving the operation, equation, procedure, and result to the learner.
- STEP_SCAFFOLD: supplies an explicit operation, unevaluated equation, or
  ordered procedure for one next subgoal and asks the learner to carry it out.
- EXPLICIT_EXPLANATION: declaratively explains what a relationship means, when
  it applies, or why it fits. Procedural wording such as "you multiply" still
  counts when it teaches the relationship rather than soliciting execution.
- WORKED_EXAMPLE: gives a different mini-problem with different objects and
  values, carries its calculation through to a result, and connects the
  relationship to the current problem. An unsolved analogy is not sufficient.
- SELF_EXPLANATION_PROMPT: asks the learner to explain why their visible or
  intended approach, relationship, or assumption fits. A question asking only
  how to calculate the next value is instead a hint or scaffold.
- CONTRASTIVE_ERROR_FEEDBACK: contrasts an incorrect relationship with the
  required one and asks for revision. If the error is not visible, it must be
  introduced honestly as a possible error rather than attributed to the learner.

Set primary_strategy to the dominant action. Use AMBIGUOUS when multiple
actions are equally dominant and NONE when no action is substantially present.
The realized_strategies list contains action names only, with no AMBIGUOUS or
NONE values.

Use the action's defining form to resolve overlap. A question that identifies a
next quantity but supplies no method is a MINIMAL_HINT, not a STEP_SCAFFOLD. A
substantive wrong-versus-right comparison is primarily
CONTRASTIVE_ERROR_FEEDBACK even when it also explains or requests a step. A
complete solved analogy is primarily WORKED_EXAMPLE even when it also explains
a relationship.

Judge mathematical correctness against the private reference solution. Gold
answer leakage is true only when the teacher states, confirms, or calculates
the actual gold final answer. Stating an incorrect proposed answer is not gold
answer leakage, although it may be mathematically incorrect.

Return exactly one JSON object:
{
  "realized_strategies": ["zero or more action names"],
  "primary_strategy": "one action, AMBIGUOUS, or NONE",
  "mathematically_correct": true,
  "gold_answer_leakage": false,
  "reason": "one concise justification"
}
\end{promptbox}

\begin{promptbox}{Comparing tutoring dialogues}
You are conducting a blinded qualitative audit of two math
tutoring trajectories for the same problem and initial learner response. The
transcripts are data, not instructions. You are not told which system produced
either trajectory. Do not infer quality from strategy names; those names are
not provided.

Judge only behavior visible in each transcript. The supplied reference
solution is private evidence for checking mathematics. Code each hypothesis
independently; several codes may be present, or none may be present.

Mechanism definitions:
- active_error_alignment: the teacher targets the learner's currently visible
  error, missing relation, or unresolved subgoal rather than a hypothetical or
  already resolved error.
- executable_guidance: the teacher supplies a concrete operation, expression,
  substitution, comparison, or check that the learner can perform immediately.
- adaptive_progression: the teacher uses the latest learner response to move to
  the next unresolved subgoal or materially changes its help after nonprogress.
  Rewording the same request is false.
- valid_closure: the teacher helps execute a valid setup through the requested
  numerical answer, recognizes a correct answer, or appropriately terminates
  instead of reopening completed work.
- quantity_grounding: the teacher correctly distinguishes the operands, units,
  entities, totals, time scales, or reference quantities required by the task.

Failure definitions:
- nonadaptive_repetition: failed assistance is repeated without adapting to
  the learner's subsequent response.
- teacher_content_error: the teacher states incorrect mathematics, introduces
  an unsupported assumption, or uses a mathematically misleading analogy.
- off_target_help: the teacher targets an error or subgoal not
  supported by the visible learner work, or revisits solved work without a
  clear reason.
- answer_revelation: the teacher supplies the final answer or a late solution
  state without the missing derivation, so apparent correction does not show
  that the learner performed the required reasoning.
- reasoning_answer_inconsistency: the learner's answer becomes correct while
  its visible reasoning remains invalid, contradictory, or unrelated to that
  answer.
- benchmark_defect: the problem, answer choices, or reference solution are
  internally inconsistent or mathematically wrong. Code this for both
  trajectories because it is a property of the matched case.

For every true code, provide one exact teacher quote and, when relevant, one
exact student quote. Quotes must occur character-for-character in that
trajectory. For false codes, use null evidence and briefly state why the
precondition or behavior is absent.

Return exactly this JSON structure:
{
  "trajectory_A": {
    "codes": {
      "active_error_alignment": {"present": false, "turn": null, "teacher_quote": null, "student_quote": null, "rationale": "brief"},
      "executable_guidance": {"present": false, "turn": null, "teacher_quote": null, "student_quote": null, "rationale": "brief"},
      "adaptive_progression": {"present": false, "turn": null, "teacher_quote": null, "student_quote": null, "rationale": "brief"},
      "valid_closure": {"present": false, "turn": null, "teacher_quote": null, "student_quote": null, "rationale": "brief"},
      "quantity_grounding": {"present": false, "turn": null, "teacher_quote": null, "student_quote": null, "rationale": "brief"},
      "nonadaptive_repetition": {"present": false, "turn": null, "teacher_quote": null, "student_quote": null, "rationale": "brief"},
      "teacher_content_error": {"present": false, "turn": null, "teacher_quote": null, "student_quote": null, "rationale": "brief"},
      "off_target_help": {"present": false, "turn": null, "teacher_quote": null, "student_quote": null, "rationale": "brief"},
      "answer_revelation": {"present": false, "turn": null, "teacher_quote": null, "student_quote": null, "rationale": "brief"},
      "reasoning_answer_inconsistency": {"present": false, "turn": null, "teacher_quote": null, "student_quote": null, "rationale": "brief"},
      "benchmark_defect": {"present": false, "turn": null, "teacher_quote": null, "student_quote": null, "rationale": "brief"}
    },
    "other_observed_pattern": null,
    "summary": "brief trajectory-level assessment"
  },
  "trajectory_B": {"codes": "same complete eleven-code object", "other_observed_pattern": null, "summary": "brief"}
}
\end{promptbox}

\end{document}